\documentclass{article} % For LaTeX2e
\usepackage{iclr2027_conference,times}
\iclrfinalcopy

\usepackage{amsmath,amsfonts,bm}

\def\eqref#1{equation~\ref{#1}}
\def\1{\bm{1}}

\DeclareMathAlphabet{\mathsfit}{\encodingdefault}{\sfdefault}{m}{sl}
\SetMathAlphabet{\mathsfit}{bold}{\encodingdefault}{\sfdefault}{bx}{n}

\usepackage{hyperref}
\usepackage{url}
\usepackage{enumitem}
\usepackage{amsmath,amssymb,amsthm}
\usepackage{algorithm}
\usepackage{algorithmic}

\newtheorem{proposition}{Proposition}
\newtheorem{corollary}[proposition]{Corollary}
\usepackage{graphicx}
\usepackage{booktabs}

\title{Prior-Data Fitted Networks for In-Context Off-Policy Evaluation}

\author{Shuguang Yu$^{a}$, Ting Li$^{b}$, and Fan Zhou$^{b}$\thanks{Corresponding author. Email: {zhoufan@mail.shufe.edu.cn}}\\
$^a$The Hong Kong Polytechnic University, Hong Kong, China\\
$^b$Shanghai University of Finance and Economics, Shanghai, China}

\begin{document}

\maketitle
\lhead{Preprint}

\begin{abstract}
Off-policy evaluation (OPE) is essential for assessing decision policies before deployment when online experimentation is costly or unsafe. Accurate estimation remains difficult when behavior-policy data are limited and the target policy favors states and actions rarely observed in those data. We propose InCOPE, a prior-data fitted network that learns a shared policy-value prediction rule across synthetic OPE tasks. We construct an OPE task prior over Markov decision processes and behavior–target policy pairs, generating behavior-policy datasets and exact target-policy values for pretraining. For a new tabular decision problem, the pretrained network uses the behavior-policy dataset, target policy, and known task settings as context to predict the policy’s expected return without task-specific parameter updates. Experiments across four tabular domains, including a semi-synthetic clinical setting, demonstrate InCOPE's effectiveness for policy-value estimation.
\end{abstract}

\section{Introduction}
Evaluating a decision policy before deployment is essential when online experimentation is costly or unsafe. This concern arises in applications such as healthcare and intelligent tutoring \citep{gao2023humanfeedback,gao2024trajectory,yu2024twoway}, online advertising \citep{xu2023instrumental}, and ridesharing \citep{shi2024confounded}. Off-policy evaluation (OPE) estimates a target policy's expected cumulative return from trajectories collected under a behavior policy, without deploying the target policy.

A substantial body of work has developed importance-sampling, model-based and value-based, doubly robust, and marginalized-ratio estimators for OPE \citep{precup2000eligibility,le2019batch,jiang2016dr,liu2018breaking}. With limited offline data, however, states and actions frequently visited by the target policy may be poorly represented, particularly under substantial behavior--target policy mismatch. Importance weighting can then suffer from high variance, while model-based and value-based methods can produce inaccurate estimates for poorly observed state--action pairs \citep{le2019batch,liu2018breaking}. These difficulties motivate using knowledge acquired from related decision problems to improve value estimation with limited current-task data.

% Pretraining offers a way to encode this knowledge in an estimator's parameters for use in new tasks. This principle underlies prior-data fitted networks (PFNs), which learn prediction rules from synthetic datasets sampled from a specified task prior \citep{muller2022pfn,nagler2023pfn}. PFNs have been applied to tabular prediction \citep{hollmann2025tabpfn}, causal inference \citep{ma2026causalfm}, and policy learning in reinforcement learning \citep{schiff2026icrrl,zighem2026rlfm}. For OPE, this suggests learning a value estimator that draws on knowledge acquired across source tasks when current-task data are scarce. The pretrained estimator would predict policy values using the new task's offline data, target policy, and known settings as context, without parameter updates.

Pretraining can encode cross-task knowledge in an estimator's parameters. Prior-data fitted networks (PFNs) learn from synthetic datasets sampled from a task prior \citep{muller2022pfn,nagler2023pfn}, with applications to tabular prediction \citep{hollmann2025tabpfn}, causal inference \citep{ma2026causalfm}, and policy learning \citep{schiff2026icrrl,zighem2026rlfm}. For OPE, however, extending PFNs to the evaluation of state-dependent target policies remains unexplored, to our knowledge. The challenge is to infer a specified target policy's cumulative return from finite behavior-policy observations, accounting for how its actions and transition dynamics jointly determine that return.

% To learn such an estimator, we introduce InCOPE (In-Context Off-Policy Evaluation), a prior-data fitted network for OPE. During pretraining, we sample Markov decision processes (MDPs), behavior policies, and target policies from a task prior, generate finite datasets under the behavior policies, and train the network to predict the exact target-policy values computed in the sampled MDPs. The network encodes state--action features such as mean rewards and visit counts, aggregates these representations according to the target policy, and propagates information using transitions estimated from the data. At test time, the frozen network returns a value estimate in a single forward pass without additional environment interactions. We focus on finite tabular MDPs, which allow exact reference policy values and controlled comparisons across task distributions.

To bridge this gap, we introduce InCOPE (In-Context Off-Policy Evaluation), a prior-data fitted network for OPE. We learn to predict target-policy returns from behavior-policy observations by pretraining on finite datasets paired with exact target-policy values, generated from Markov decision processes (MDPs) and policy pairs sampled from a task prior. The network encodes mean rewards and visit counts for each state--action pair, then combines target-policy-weighted aggregation with repeated propagation through estimated transitions. This incorporates successor-state information under the queried policy, connecting observed local statistics to multi-step returns. At test time, the frozen network returns a value estimate in a single forward pass without additional environment interactions. We focus on finite tabular MDPs, which allow exact reference policy values and controlled comparisons across task distributions.

We summarize our contributions as follows.
\begin{itemize}[leftmargin=*]
    \item We introduce InCOPE, a prior-data fitted network that learns a shared policy-value prediction rule across synthetic OPE tasks. The pretrained network estimates target-policy values in new tasks from historical data collected under behavior policies, without task-specific parameter updates.

    \item We formulate a joint prior over MDPs, behavior and target policies, and observed trajectories, and characterize the Bayesian policy-value predictor targeted by prior fitting.

    \item Our empirical results across four structurally distinct tabular OPE task families show that InCOPE improves policy-value estimation relative to representative OPE estimators when behavior-policy data are limited and behavior and target policies differ substantially.
\end{itemize}

\section{Related Work}

\paragraph{OPE estimators.}
Existing OPE estimators use prediction, reweighting, or both. Value-based methods such as fitted Q evaluation learn target-policy value functions through Bellman regression \citep{le2019batch}, whereas model-based methods evaluate the target policy using learned reward and transition models \citep{gottesman2019combining}. Importance sampling instead reweights observed returns using behavior- and target-policy action probabilities \citep{precup2000eligibility}. Marginalized estimators and density-ratio methods replace products of importance ratios along trajectories with state or state--action distribution ratios \citep{liu2018breaking,yin2020asymptotically,nachum2019dualdice,uehara2020minimax}. Doubly robust estimators bring prediction and reweighting together by combining value estimates with importance-weighted residual corrections \citep{jiang2016dr,thomas2016dataefficient}. Further refinements optimize regression-based corrections for statistical efficiency \citep{farajtabar2018mrdr,kallus2019intrinsically} or use reward shaping to construct control variates for sparse rewards \citep{majumdar2026reward}. At the estimator level, OPERA uses bootstrap estimates of joint errors to combine multiple candidate estimators \citep{nie2024opera}. These studies improve estimation within individual OPE tasks, but do not address how to learn a reusable value-prediction rule from related MDPs when data from a new task are scarce.

\paragraph{OPE with auxiliary information.}
Beyond estimator design, structural priors and trajectory augmentation can improve OPE with limited data. Structured Bayesian OPE uses an informative prior to share reward information across correlated actions in contextual bandits \citep{aouali2024bayesian}. For sequential problems, OAT mines and augments sub-trajectories from behavior-policy data for downstream OPE \citep{gao2024trajectory}. STITCH-OPE uses policy-guided diffusion to generate and stitch target-policy trajectories, then evaluates policies using their predicted returns \citep{goli2025stitch}. Auxiliary rollouts can also be combined with observed behavior-policy data, as in PERRY, which uses correction or calibration to obtain value estimates and confidence intervals and permits pretrained generators \citep{mandyam2026perry}. A complementary route encodes cross-task knowledge directly in a pretrained value predictor without generating extra trajectories at inference time.

\paragraph{Prior-data fitted networks.}
PFNs learn reusable predictors from synthetic tasks sampled from a prior \citep{muller2022pfn,nagler2023pfn}. TabPFN applies this approach to tabular classification and regression \citep{hollmann2023tabpfn,hollmann2025tabpfn}; LimiX-2 demonstrates synthetic pretraining's potential for tabular prediction \citep{zhang2026limix2}. CausalPFN, Do-PFN, and CausalFM extend this approach to causal inference \citep{balazadeh2025causalpfn,robertson2025dopfn,ma2026causalfm}. In reinforcement learning, ICR-RL and PFN-RL use PFNs for policy learning \citep{schiff2026icrrl,zighem2026rlfm}. For large language models, $V_0$ uses a TabPFN head to predict a model's success probability on a new prompt from that model's observed performance on previous prompts \citep{zhang2026v0}. Sequential OPE poses a different prediction problem, requiring the expected return of a specified target policy to be inferred from limited data collected under another policy in a new MDP. InCOPE addresses this setting by pretraining a shared value predictor on synthetic MDPs, with behavior-policy data as context and exact target-policy values as supervision.

\section{Problem Formulation}
\label{sec:problem-background}

In this section, we formulate the OPE problem and introduce the notation used throughout the paper.

Consider a finite-horizon, time-homogeneous MDP $\mathcal M=(\mathcal S,\mathcal A,P,r,\rho_0,\gamma,H)$. Here $\mathcal S$ and $\mathcal A$ are finite state and action
spaces, $P(s'\mid s,a)$ is the transition probability, and $r(\cdot\mid s,a)$ is the reward distribution with finite mean. The initial-state distribution is $\rho_0$, $\gamma\in(0,1]$ is the discount factor, and $H\geq1$ is an integer horizon. We assume that the reward and next state are independent given $(s,a)$.

The behavior policy $\pi_b$ and target policy $\pi_e$ are stationary Markov policies, with $\pi(a\mid s)$ denoting the probability of choosing action $a$ in state $s$ under policy $\pi$. A trajectory is $\tau=(S_0,A_0,R_0,\ldots,S_H)$, where $S_t$,
$A_t$, and $R_t$ denote the state, action, and reward at step $t$, and $S_0\sim\rho_0$.
For fixed $\mathcal M$ and $\pi_b$, the dataset $D_n=(\tau_1,\ldots,\tau_n)$ contains $n$ i.i.d. trajectories generated under $\pi_b$. The usual coverage condition requires that, over the $H$-step horizon, every state--action pair visited with positive probability under $\pi_e$ also has positive visitation probability under
$\pi_b$ \citep{precup2000eligibility,thomas2016dataefficient,yin2020asymptotically}.
Even under this condition, a finite dataset may leave some target-relevant state--action pairs unobserved.

The value of the target policy $\pi_e$ in $\mathcal M$ is its expected discounted return
\begin{equation}
  J(\mathcal M,\pi_e)
  :=\mathbb E_{\mathcal M,\pi_e}
  \!\left[\sum_{t=0}^{H-1}\gamma^tR_t\right].
  \label{eq:ope-estimand}
\end{equation}
OPE aims to estimate $J(\mathcal M,\pi_e)$ from $D_n$ without executing $\pi_e$ in the environment \citep{precup2000eligibility,jiang2016dr}. In our setting, the task specification $(\mathcal S,\mathcal A,\rho_0,\gamma,H)$ and target policy $\pi_e$ are known, whereas $P$, $r$, and $\pi_b$ are unknown.

% \subsection{Prior-Data Fitted Inference}
% \label{sec:prior-fitted-inference}

% Prior-data fitted networks (PFNs) are pretrained on synthetic tasks to predict labels from a context dataset and a query \citep{muller2022pfn,nagler2023pfn}. Let $\mathcal T\sim\Pi$ denote a supervised learning task drawn from a generative prior $\Pi$. A labeled context dataset $\mathcal D$ and a held-out input--label pair $(x,y)$ are generated from this task. The network $q_\theta(\cdot\mid x,\mathcal D)$, with parameters
% $\theta$, takes $\mathcal D$ and the query $x$ as input and predicts a distribution over $y$.

% The population pretraining objective is the expected negative log-likelihood
% \begin{equation*}
%   \mathcal L_\Pi(\theta)
%   :=\mathbb E_{\mathcal T\sim\Pi}
%   \mathbb E_{\mathcal D,x,y\mid\mathcal T}
%   \!\left[-\log q_\theta(y\mid x,\mathcal D)\right].
% \end{equation*}
% Training approximates this expectation by averaging over simulated datasets and held-out examples. This objective targets the posterior predictive distribution under $\Pi$ and equals the expected Kullback--Leibler divergence from that distribution to $q_\theta$, up to a constant independent of $\theta$. With fitted parameters $\widehat\theta$ fixed, the network returns $q_{\widehat\theta}(\cdot\mid x,\mathcal D)$ for a new dataset $\mathcal D$ and query $x$.

\section{InCOPE: In-Context Off-Policy Evaluation}
\label{sec:incope}

InCOPE is pretrained on synthetic OPE tasks with known policy values. Given a target policy and a new dataset collected under a behavior policy, it returns a value estimate in a forward pass without parameter updates. This section describes task priors, network design, pretraining, and evaluation.

\subsection{Prior-Fitted Policy-Value Estimation}
\label{sec:ope-task-prior}

Given a behavior-policy dataset $D_n$ and a target policy $\pi_e$, we construct a context $C_n$ by deterministic preprocessing of these inputs, the known task settings $(\rho_0,\gamma,H)$, and any supplied action attributes. Let $f_\theta$ denote InCOPE's neural predictor, with $\theta$ collecting all its trainable parameters. We denote its estimate of $J(\mathcal M,\pi_e)$ by
\begin{equation}
  \widehat J_\theta:=f_\theta(C_n).
  \label{eq:incope-estimator}
\end{equation}
The parameters are learned across source OPE tasks and fixed at $\widehat\theta$ when evaluating new tasks. The context representation is specified in Eq.~\eqref{eq:incope-context-map} in Section~\ref{sec:incope-representation}.

For pretraining, we define a source OPE task as $\Xi:=(\mathcal M,n,\pi_b,\pi_e)$ and specify a joint task prior $\Pi_{\mathrm{pre}}$. We sample tasks and datasets according to
\begin{equation}
  \Xi\sim\Pi_{\mathrm{pre}},
  \qquad
  D_n\sim p(\,\cdot\mid\mathcal M,n,\pi_b),
  \label{eq:incope-task-record}
\end{equation}
where $p(\,\cdot\mid\mathcal M,n,\pi_b)$ is the distribution of $n$ independent trajectories generated under $\pi_b$ in $\mathcal M$. The task prior allows $n$, $H$, $\pi_b$, and $\pi_e$ to vary during pretraining. This supports in-context adaptation to different data budgets, horizons, and policy pairs without task-specific parameter updates. The supervision is the exact value $J(\mathcal M,\pi_e)$, computed by finite-horizon dynamic programming. For notational simplicity, we use $J$ as shorthand for $J(\mathcal M,\pi_e)$ hereafter. The true dynamics and reward distributions are available to the source-task generator but not to the predictor.

For multiple source task families, let $\mathcal F$ be their finite index set and $F\in\mathcal F$ an individual family index. The family-specific prior $\Pi_F$ is a distribution over $\Xi$, determined by its MDP generator, policy-construction rules, and trajectory-count distribution. We define $\Pi_{\mathrm{pre}}:=\sum_{F\in\mathcal F}w_F\Pi_F$, where $w_F\geq0$ is the probability of selecting family $F$ and $\sum_{F\in\mathcal F}w_F=1$. These weights can reflect anticipated task-family frequencies or application priorities. We use uniform weights, $w_F=1/|\mathcal F|$, in our experiments. The population pretraining objective is
\begin{equation}
  \mathcal L_{\mathrm{pre}}(\theta)
  :=\sum_{F\in\mathcal F}w_F\,
    \mathbb E_{\Xi\sim\Pi_F}
    \mathbb E_{D_n\mid\Xi}
    \left[
      \frac{\bigl(f_\theta(C_n)-J\bigr)^2}{s_F^2}
    \right],
  \label{eq:incope-population-objective}
\end{equation}
where $s_F>0$ measures the typical spread of family $F$'s source-training target-policy values and is held fixed. Dividing by $s_F^2$ expresses squared errors relative to this spread; without it, larger-scale families can dominate the objective even at comparable relative errors. The two expectations average over source tasks and their behavior-policy datasets, respectively, but do not remove these scale differences. See Appendices~\ref{app:incope-training} and~\ref{app:task-priors} for scale estimation and family sampling.

The fixed loss scales reweight the families, giving the effective task prior $\Pi_{\mathrm{eff}}:=(\sum_{F\in\mathcal F}w_Fs_F^{-2}\Pi_F)/(\sum_{F\in\mathcal F}w_Fs_F^{-2})$. We write $\mathbb E_{\mathrm{eff}}$ for expectation over tasks $\Xi\sim\Pi_{\mathrm{eff}}$ and datasets $D_n\sim p(\,\cdot\mid\mathcal M,n,\pi_b)$ generated conditionally on those tasks. For a specialist, or when all family scales are equal, $\Pi_{\mathrm{eff}}=\Pi_{\mathrm{pre}}$. The population objective is a positive constant multiple of squared risk under $\Pi_{\mathrm{eff}}$ and therefore has the same minimizers.

\begin{proposition}[Bayes-optimal policy-value prediction]
\label{prop:incope-bayes}
Suppose $\mathbb E_{\mathrm{eff}}[J^2]<\infty$. Consider the objective in Eq.~\eqref{eq:incope-population-objective} with $f_\theta$ replaced by an arbitrary measurable predictor $f$ satisfying $\mathbb E_{\mathrm{eff}}[f(C_n)^2]<\infty$. This objective admits a minimizer, denoted by $f_{\mathrm{pre}}^\star$, that is unique up to sets of contexts with zero probability under effective pretraining and satisfies
\[
  f_{\mathrm{pre}}^\star(C_n)
  =\mathbb E_{\mathrm{eff}}[J\mid C_n]
\]
almost surely under effective pretraining.
\end{proposition}

Proposition~\ref{prop:incope-bayes} identifies the predictor based on $C_n$ that minimizes expected squared error under the effective prior, using source-task knowledge without uniquely identifying the MDP. Under the conditions of Corollary~\ref{cor:incope-compatibility}, this ideal predictor also minimizes test squared error when effective-pretraining and test conditional means of $J$ given $C_n$ agree. Both proofs appear in Appendix~\ref{app:incope-theory}.

\begin{corollary}[Conditional-mean compatibility]
\label{cor:incope-compatibility}
Under the assumptions of Proposition~\ref{prop:incope-bayes}, let $\mathbb E_{\mathrm{test}}$ denote expectation over test tasks and their datasets. Suppose $\mathbb E_{\mathrm{test}}[J^2]<\infty$, every measurable context set with zero effective-pretraining probability also has zero test probability, and $\mathbb E_{\mathrm{test}}[J\mid C_n]=\mathbb E_{\mathrm{eff}}[J\mid C_n]$ almost surely under the test distribution. Then $f_{\mathrm{pre}}^\star$ minimizes $\mathbb E_{\mathrm{test}}[(f(C_n)-J)^2]$ over measurable predictors $f$ with $\mathbb E_{\mathrm{test}}[f(C_n)^2]<\infty$.
\end{corollary}

\subsection{Network Architecture and Training}
\label{sec:incope-representation}

\begin{figure}[htbp]
  \vskip -0.1in
  \centering
  \includegraphics[width=\linewidth]{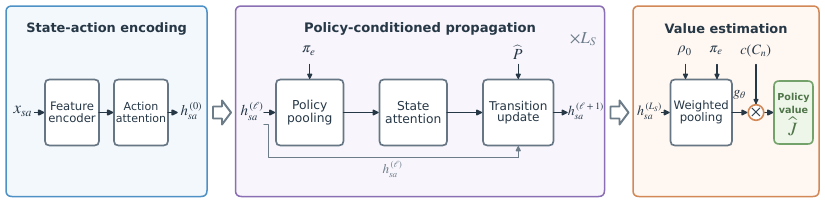}
  \vskip -0.15in
  \caption{Network architecture of InCOPE.}
  \label{fig:incope-architecture}
  \vskip -0.2in
\end{figure}

Given the task prior and objective, we specify the context $C_n$ and a Transformer predictor $f_\theta$ with state--action encoding, policy-conditioned propagation, and value estimation (Figure~\ref{fig:incope-architecture}).

\paragraph{State--action encoding.}
We organize statistics from $D_n$ by state--action pair, as in PFN-RL \citep{zighem2026rlfm}. For each $(s,a)$, a numerical feature vector $\mathbf x_{sa}$ records visit counts and reward statistics, together with target-policy probabilities, behavior-policy probabilities estimated from $D_n$, and known task information. Let $\eta:=(\rho_0,\gamma,H,\mathbf d_A)$ collect the known task settings, where $\mathbf d_A:=(\mathbf d_a)_{a\in\mathcal A}$ contains any supplied action-attribute vectors, such as treatment dose levels. A deterministic preprocessing map $\phi$ constructs the context
\begin{equation}
  C_n:=\phi(D_n,\pi_e;\eta)
    =\bigl((\mathbf x_{sa})_{s\in\mathcal S,a\in\mathcal A},\widehat P\bigr),
  \label{eq:incope-context-map}
\end{equation}
where $\widehat P$ is the smoothed transition tensor estimated from $D_n$. The target policy and the quantities in $\eta$ are included in the features $\mathbf x_{sa}$, while $\widehat P$ is retained separately. A shared multilayer perceptron serves as the feature encoder, mapping $\mathbf x_{sa}$ to hidden vectors following the numerical-input encoding used in PFNs \citep{muller2022pfn,hollmann2023tabpfn}. Action attention operates across actions within each state and produces an initial hidden representation $h_{sa}^{(0)}$ for each state--action pair. The feature definitions and preprocessing rules are given in Appendix~\ref{app:incope-input-features}.

\paragraph{Policy-conditioned propagation.}
The initial representations pass through $L_S$ propagation blocks, each using policy pooling to form target-policy-weighted state representations, state attention to exchange information across states, and a transition update to incorporate successor information. Let $h_{sa}^{(\ell)}$ denote the representation after $\ell$ blocks. For $\ell=0,\ldots,L_S-1$, we compute
\begin{equation}
  \begin{aligned}
    u_s^{(\ell)}
      &:=\sum_a\pi_e(a\mid s)h_{sa}^{(\ell)},
    &\widetilde u^{(\ell)}
      &:=\operatorname{Attn}_{S,\ell}(u^{(\ell)}),\\
    m_{sa}^{(\ell)}
      &:=\sum_{s'}\widehat P(s'\mid s,a)\widetilde u_{s'}^{(\ell)},
    &h_{sa}^{(\ell+1)}
      &:=\Psi_\ell\!\left(h_{sa}^{(\ell)},
        \widetilde u_s^{(\ell)},m_{sa}^{(\ell)},H\right),
  \end{aligned}
  \label{eq:incope-transition-message}
\end{equation}
where $u^{(\ell)}:=(u_s^{(\ell)})_{s\in\mathcal S}$ and $\widetilde u^{(\ell)}:=(\widetilde u_s^{(\ell)})_{s\in\mathcal S}$ collect state vectors before and after state attention $\operatorname{Attn}_{S,\ell}$. The transition update computes $m_{sa}^{(\ell)}$ using $\widehat P$. The learnable map $\Psi_\ell$ then combines this successor message with the retained state--action and updated current-state vectors, conditioned on the known horizon $H$. Each state--action pair retains a separate representation and receives successor information through its own empirical transition distribution. The attention blocks and update map, including horizon encoding and gating, are detailed in Appendix~\ref{app:incope-network}.

\paragraph{Value estimation.}
Weighted pooling averages the final representations over actions under $\pi_e$ and over states under $\rho_0$. A scalar prediction head and a context-dependent output scale then give
\begin{equation}
  f_\theta(C_n)
    :=c(C_n)\,g_\theta\!\left(
      \sum_s\rho_0(s)\sum_a\pi_e(a\mid s)h_{sa}^{(L_S)}
    \right),
  \label{eq:incope-readout}
\end{equation}
where $g_\theta$ is a multilayer perceptron with a scalar output and $c(C_n)>0$ is a deterministic output scale computed from the current dataset's reward statistics and $\sum_{t=0}^{H-1}\gamma^t$. The fixed scales $s_F$ normalize the training loss, whereas $c(C_n)$ sets the output scale for the current task. The formula for $c(C_n)$ is given in Appendix~\ref{app:incope-input-features}, and the prediction head and padding rules are specified in Appendix~\ref{app:incope-network}.

\paragraph{Training.}
We fit the network by minimizing the empirical counterpart of Eq.~\eqref{eq:incope-population-objective}. For each family $F$, let $\{(C_{Fj},J_{Fj})\}_{j=1}^{M_F}$ denote its source-training queries, with $C_{Fj}$ and $J_{Fj}$ respectively giving the context and exact target-policy value of query $j$. The empirical objective is
\begin{equation}
  \widehat{\mathcal L}_{\mathrm{pre}}(\theta)
    :=\sum_{F\in\mathcal F}\frac{w_F}{M_F}
      \sum_{j=1}^{M_F}
      \left(\frac{f_\theta(C_{Fj})-J_{Fj}}{s_F}\right)^2,
  \label{eq:incope-empirical-objective}
\end{equation}
where $M_F$ counts source-training queries in family $F$, whereas $n$ counts trajectories within each query. For the main comparison, each training replicate uses 3,000 training and 600 validation queries per family, generated before optimization and held fixed. Source MDP groups are disjoint between training and validation. We select $\widehat\theta$ by source-validation loss. Training and evaluation are summarized in Algorithm~\ref{alg:incope}, with optimization and model-selection settings in Appendix~\ref{app:incope-training}.

\vskip -0.1in
\begin{algorithm}[htbp]
  \caption{Training and evaluation of InCOPE}
  \label{alg:incope}
  \begin{algorithmic}[1]
    \REQUIRE Family priors $\{\Pi_F\}$, weights $\{w_F\}$, and source-set sizes;
      new dataset $D_n$, target policy $\pi_e$, and known inputs $\eta$
    \STATE \textbf{Training}
    \STATE Generate training/validation tasks and data via Eq.~(\ref{eq:incope-task-record}), with disjoint MDP groups
    \FOR{each generated task $\Xi=(\mathcal M,n,\pi_b,\pi_e)$ and its dataset $D_n$}
      \STATE Construct $C_n\gets\phi(D_n,\pi_e;\eta)$ with this task's known $\eta$ via Eq.~(\ref{eq:incope-context-map})
      \STATE Compute the label $J$ by finite-horizon dynamic programming in $\mathcal M$ under $\pi_e$
    \ENDFOR
    \STATE Fix each $s_F$ using its family's training labels
    \STATE Optimize Eq.~\eqref{eq:incope-empirical-objective}; select $\widehat\theta$ by source-validation loss
    \STATE \textbf{Evaluation}
    \STATE For the new task, construct $C_n\gets\phi(D_n,\pi_e;\eta)$ and return $\widehat J_{\widehat\theta}:=f_{\widehat\theta}(C_n)$ with $\widehat\theta$ fixed
  \end{algorithmic}
\end{algorithm}
\vskip -0.15in

\section{Experiments}
\label{sec:experiments}

In this section, we evaluate whether pretraining across related OPE tasks improves policy-value estimation from limited behavior-policy data. Comparisons across four tabular domains assess accuracy under varying data sizes and behavior--target policy mismatch, while controlled studies examine the roles of the pretraining prior, current-task observations, and estimator design.

\subsection{Experimental Setup}
\label{sec:experimental-setup}

We evaluate InCOPE in four finite tabular domains. GridWorld models stochastic navigation with grid-cell states, four directional actions, and varying layouts and reward mechanisms. Tumor models treatment decisions in discretized tumor-growth systems, where chemotherapy and radiotherapy affect tumor burden and incur treatment penalties. Dispatch represents a vehicle-service market, with states describing the market zone and supply--demand imbalance, and actions combining driver incentives and passenger promotions. The semi-synthetic Clinical evaluation uses the ICU-Sepsis MDP \citep{choudhary2024icusepsis}, with aggregated patient states and 25 fluid--vasopressor treatment combinations. Pretraining and evaluation use distinct environment generators within each domain. Dynamics, rewards, and policy construction are detailed in Appendix~\ref{app:task-priors}.

We evaluate two InCOPE configurations: domain-specific specialists pretrained on synthetic tasks from their respective domains, and a unified model pretrained on pooled tasks from all four domains. The unified model has the same architecture and parameter count as each specialist and uses no domain identifier. Both configurations are evaluated in Sections~\ref{sec:ope-performance} and~\ref{sec:log-policy-sensitivity}; other experiments use specialists. We train each configuration independently five times and select checkpoints using source-validation data. Training details are given in Appendix~\ref{app:incope-training}.

We compare with six OPE estimators. Weighted per-decision importance sampling (WPDIS) \citep{precup2000eligibility} reweights observed rewards using self-normalized importance weights. Fitted Q evaluation (FQE) \citep{le2019batch} fits action-value functions by Bellman regression. Tabular marginalized importance sampling (Tabular-MIS) \citep{yin2020asymptotically} estimates target-policy state occupancies and combines them with estimated rewards. Model-based OPE (MB-OPE) \citep{jiang2016dr} fits a pooled empirical MDP and evaluates the target policy by dynamic programming. OPERA \citep{nie2024opera} combines candidate OPE estimators using data-dependent weights. Our tabular implementation of PERRY's DR-PPI \citep{mandyam2026perry} corrects model-based value estimates using held-out importance-sampling returns. See Appendix~\ref{app:baseline-implementations} for implementation details.

In baseline comparisons, all methods share current-task data and target policies, with required behavior-policy probabilities estimated from the data. InCOPE is pretrained with the loss in Eq.~(\ref{eq:incope-population-objective}) and frozen at test time. We measure mean absolute error (MAE) in original return units against exact target-policy values from finite-horizon dynamic programming. Tables~\ref{tab:incope-prior-ablations} and~\ref{tab:incope-information-estimator-controls} report mean MAE $\pm$ sample SD over five training replicates, with a separate shared test set per panel.

\subsection{Comparison with Existing OPE Estimators}
\label{sec:ope-performance}

We first examine whether a prior-fitted predictor can improve policy-value estimation from limited behavior-policy data. Table~\ref{tab:incope-main-results} reports results for 6,700 test dataset--policy pairs, with the number of trajectories per dataset, $n$, varying over the domain-specific ranges shown in the table. Each domain contains five separately generated test batches. For InCOPE, each batch-level score is the average MAE of the five pretrained models. For every method, we report the mean and sample standard deviation of these batch-level scores. The full evaluation budgets are given in Appendix~\ref{app:main-evaluation-details}.

% \begin{table}[htbp]
%     \vskip -0.1in
%     \centering
%     \caption{Comparison of InCOPE with existing OPE estimators across four domains.}
%     \label{tab:incope-main-results}
%     \small
%     \setlength{\tabcolsep}{3pt}
%     \begin{tabular}{@{}lcccc@{}}
%         \toprule
%         Method & GridWorld & Tumor & Dispatch & Clinical \\
%         & $n=4$--$32$ & $n=8$--$48$ & $n=4$--$24$ & $n=24$--$160$ \\
%         \midrule
%         WPDIS & $1.519\pm0.081$ & $1.229\pm0.062$ & $0.566\pm0.053$ & $0.199\pm0.008$ \\
%         FQE & $0.982\pm0.065$ & $2.329\pm0.030$ & $13.248\pm0.192$ & $0.467\pm0.008$ \\
%         Tabular-MIS & $0.982\pm0.065$ & $2.293\pm0.030$ & $13.069\pm0.199$ & $0.461\pm0.008$ \\
%         MB-OPE & $0.953\pm0.092$ & $1.686\pm0.042$ & $1.368\pm0.117$ & $0.211\pm0.017$ \\
%         OPERA & $0.985\pm0.068$ & $2.279\pm0.075$ & $13.200\pm0.200$ & $0.482\pm0.009$ \\
%         DR-PPI & $0.990\pm0.074$ & $2.083\pm0.044$ & $8.578\pm0.240$ & $0.361\pm0.008$ \\
%         \midrule
%         InCOPE (specialist) & $\mathbf{0.765}\pm0.065$ & $0.682\pm0.013$ & $\mathbf{0.318}\pm0.004$ & $\mathbf{0.065}\pm0.005$ \\
%         InCOPE (unified) & $0.766\pm0.059$ & $\mathbf{0.510}\pm0.013$ & $0.449\pm0.007$ & $0.087\pm0.006$ \\
%         \bottomrule
%     \end{tabular}
%     % \vskip -0.1in
% \end{table}

The domain-specific specialists and the unified InCOPE model both achieve lower mean MAE than every evaluated baseline in all four domains. Relative to the lowest baseline MAE in each domain, the specialists reduce error by 19.7\%--67.1\%. This advantage persists against MB-OPE, which shares InCOPE's pooled reward and transition estimates. Across these distinct decision problems, the results support using a prior-fitted estimator to improve OPE from small datasets, with source-task knowledge incorporated into a predictor that remains fixed during evaluation.

The unified model outperforms all baselines across the four pretraining domains. Compared with separate specialists, it improves on Tumor, performs similarly on GridWorld, and has higher errors on Dispatch and Clinical. Thus, a shared predictor supports OPE across these domains, although domain-specific pretraining remains preferable in some cases.

\begin{table}[htbp]
    \vskip -0.1in
    \centering
    \caption{Comparison of InCOPE with existing OPE estimators across four domains.}
    \label{tab:incope-main-results}
    \small
    \setlength{\tabcolsep}{3pt}
    \begin{tabular}{@{}lcccc@{}}
        \toprule
        Method & GridWorld & Tumor & Dispatch & Clinical \\
        & $n=4$--$32$ & $n=8$--$48$ & $n=4$--$24$ & $n=24$--$160$ \\
        \midrule
        WPDIS & $1.519\pm0.081$ & $1.229\pm0.062$ & $0.566\pm0.053$ & $0.199\pm0.008$ \\
        FQE & $0.982\pm0.065$ & $2.329\pm0.030$ & $13.248\pm0.192$ & $0.467\pm0.008$ \\
        Tabular-MIS & $0.982\pm0.065$ & $2.293\pm0.030$ & $13.069\pm0.199$ & $0.461\pm0.008$ \\
        MB-OPE & $0.953\pm0.092$ & $1.686\pm0.042$ & $1.368\pm0.117$ & $0.211\pm0.017$ \\
        OPERA & $0.985\pm0.068$ & $2.279\pm0.075$ & $13.200\pm0.200$ & $0.482\pm0.009$ \\
        DR-PPI & $0.990\pm0.074$ & $2.083\pm0.044$ & $8.578\pm0.240$ & $0.361\pm0.008$ \\
        \midrule
        InCOPE (specialist) & $\mathbf{0.765}\pm0.065$ & $0.682\pm0.013$ & $\mathbf{0.318}\pm0.004$ & $\mathbf{0.065}\pm0.005$ \\
        InCOPE (unified) & $0.766\pm0.059$ & $\mathbf{0.510}\pm0.013$ & $0.449\pm0.007$ & $0.087\pm0.006$ \\
        \bottomrule
    \end{tabular}
    % \vskip -0.1in
\end{table}

\subsection{Data Size and Policy Mismatch}
\label{sec:log-policy-sensitivity}

\begin{figure}[htbp]
    \vskip -0.15in
    \centering
    \includegraphics[width=0.9\linewidth]{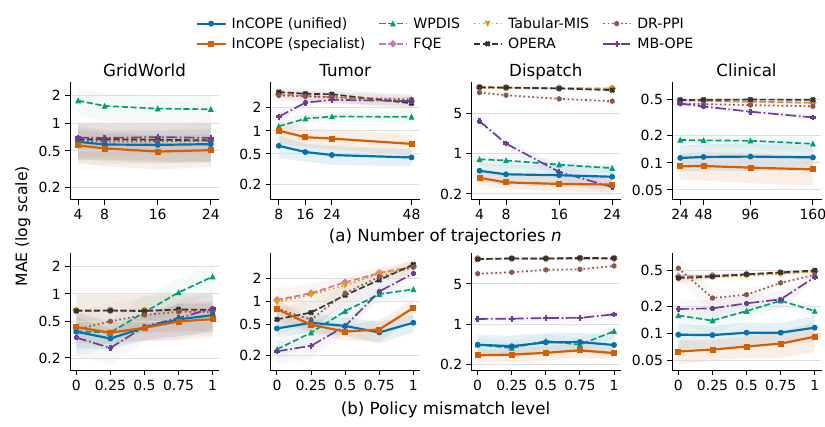}
    \vskip -0.2in
    \caption{OPE accuracy across trajectory counts and behavior--target policy mismatch levels.}
    \label{fig:incope-log-policy-shift}
    \vskip -0.2in
\end{figure}

To examine how these gains depend on evaluation conditions, we vary the amount of offline behavior-policy data and the behavior--target policy mismatch while holding each MDP and target policy fixed. Figure~\ref{fig:incope-log-policy-shift}(a) varies the number $n$ of trajectories using prefixes of the same dataset at the largest tested mismatch. Panel~(b) varies the behavior policy by interpolating from the target policy ($\alpha=0$) to the most distant candidate behavior policy ($\alpha=1$), using $n=8,16,8,48$ for GridWorld, Tumor, Dispatch, and Clinical, respectively. Shading denotes pointwise $95\%$ hierarchical-bootstrap confidence intervals; vertical axes use logarithmic scales shared across rows within each domain. The policy construction and evaluation protocol are detailed in Appendix~\ref{app:log-policy-sensitivity}.

\paragraph{Data size.}
Both InCOPE configurations achieve lower mean MAE than all baselines at the smallest tested data size in each domain, and maintain this advantage across all tested sizes in GridWorld, Tumor, and Clinical. In Dispatch, however, MB-OPE improves as more data become available for estimating the empirical MDP, reaching an MAE close to that of the specialist at $n=24$. Whereas MB-OPE estimates its reward and transition models from current-task data alone, InCOPE can also draw on knowledge acquired across source tasks. Its advantage at small trajectory counts therefore supports pretraining as a way to improve OPE when behavior-policy data are scarce.

\paragraph{Policy mismatch.}
With trajectory count held fixed, InCOPE's comparative advantage becomes clearer as the behavior and target policies diverge in GridWorld and Tumor. At the highest tested mismatch level ($\alpha=1$), both the specialist and unified models achieve lower mean MAE than all baselines in all four domains. The specialists also retain this advantage at every tested level in Dispatch and Clinical, as does the unified model in Clinical. These results support pretraining as a way to improve OPE with limited current-task data under substantial policy mismatch.

\subsection{Effect of the Pretraining Prior}
\label{sec:prior-design}

The preceding comparisons evaluate fixed pretrained models under different test conditions. We now vary source rewards and policy range to assess how pretraining tasks affect accuracy. Within each comparison, we hold the specialist architecture and source-query budget fixed and evaluate all configurations on the same test queries. Detailed protocols are given in Appendix~\ref{app:prior-ablations}.

\paragraph{Source rewards.}
We vary the proportion of source tasks retaining their original rewards ($0\%$, $50\%$, $100\%$; Table~\ref{tab:incope-prior-ablations}(a)). The remaining tasks retain their transitions and policies, but rewards at nonabsorbing steps are replaced by independent Gaussian draws with a common mean and variance within each domain. Reward-dependent inputs and exact value labels are then recomputed.

\begin{table}[htbp]
    \vskip -0.1in
    \centering
    \caption{Effects of source rewards and source-policy range on InCOPE specialists.}
    \label{tab:incope-prior-ablations}
    \small
    \setlength{\tabcolsep}{4pt}
    \begin{tabular}{@{}lcccc@{}}
        \toprule
        Setting & GridWorld & Tumor & Dispatch & Clinical \\
        \midrule
        \multicolumn{5}{@{}l}{\textit{(a) Fraction of original-reward source tasks}} \\
        $0\%$ & $1.679\pm0.094$ & $3.829\pm0.106$ & $1.718\pm1.132$ & $0.1387\pm0.0022$ \\
        $50\%$ & $0.841\pm0.042$ & $1.468\pm0.383$ & $0.336\pm0.118$ & $0.0962\pm0.0367$ \\
        $100\%$ & $\mathbf{0.764}\pm0.040$ & $\mathbf{0.660}\pm0.191$ & $\mathbf{0.318}\pm0.070$ & $\mathbf{0.0663}\pm0.0126$ \\
        \midrule
        \multicolumn{5}{@{}l}{\textit{(b) Source-policy range with original rewards fixed}} \\
        Narrow & $0.995\pm0.043$ & $1.380\pm0.625$ & $0.338\pm0.083$ & $0.0982\pm0.0286$ \\
        Broad & $\mathbf{0.769}\pm0.039$ & $\mathbf{0.686}\pm0.192$ & $\mathbf{0.321}\pm0.072$ & $\mathbf{0.0652}\pm0.0107$ \\
        \bottomrule
    \end{tabular}
    % \vskip -0.1in
\end{table}

Mean test MAE decreases as the proportion of original-reward tasks increases in every domain. Unlike the randomized alternative, the original source tasks preserve task-dependent reward patterns that inform target-policy values. Because the intervention changes both reward-dependent inputs and exact value labels, it tests the usefulness of the resulting pretraining tasks as a whole. These results indicate that, at the same source-query budget, retaining the original reward mechanisms yields better downstream estimates than the tested randomized-reward alternative.

\paragraph{Source policy range.}
With the original source rewards fixed, we next compare the original policy generator with a narrowed version that mixes both behavior and target policies with the uniform policy (Table~\ref{tab:incope-prior-ablations}(b)). Trajectories and exact value labels are regenerated for the modified policies, while the source MDPs and test distribution remain unchanged.

The original, broader policy generator yields lower mean MAE in all four domains. Since behavior policies determine the observed data and target policies specify the values to be predicted, changing both alters the OPE problems encountered during pretraining. The broader generator exposes the estimator to a more varied collection of these problems and improves downstream accuracy relative to the tested narrowing. Together with the reward comparison, this result shows that the content of source tasks matters for prior-fitted OPE even when the amount of synthetic training data is fixed.

\subsection{Current-Task Information and Estimator Design}
\label{sec:behavior-log-role}

Whereas the preceding experiments vary source tasks, we retain the original source-task supervision to examine current-task information and estimator design. We compare full InCOPE with a retrained log-free model to assess the contribution of current-task observations, and with regressors using target policies alone or complete task inputs to examine alternative predictors. All comparisons use domain-specific models and five source-training replicates. Details are given in Appendix~\ref{app:log-policy-controls}.

\paragraph{Retraining without current-task observations.}
We retrain InCOPE with all observation-derived inputs replaced by observation-independent defaults, retaining the target policy and known task specification. Source tasks, value labels, and training settings remain unchanged. The same restriction applies during training and evaluation, so training uses only inputs available at test time.

Removing these observation-derived inputs increases mean MAE in all four domains (Table~\ref{tab:incope-information-estimator-controls}(a)), with larger differences on Tumor and Dispatch. Because the restricted model is retrained with the same architecture and supervision, the comparison directly tests the contribution of current-task observations within the prior-fitted framework. These observations improve value estimation beyond what the learned prior, target policy, and known task specification provide on their own.

\begin{table}[htbp]
    \vskip -0.15in
    \centering
    \caption{Effects of current-task observations and estimator design on OPE accuracy.}
    \label{tab:incope-information-estimator-controls}
    \small
    \setlength{\tabcolsep}{4pt}
    \begin{tabular}{@{}lcccc@{}}
        \toprule
        Method & GridWorld & Tumor & Dispatch & Clinical \\
        \midrule
        \multicolumn{5}{@{}l}{\textit{(a) Retraining without log-derived information}} \\
        Full InCOPE & $\mathbf{0.785}\pm0.045$ & $\mathbf{0.680}\pm0.190$ & $\mathbf{0.323}\pm0.069$ & $\mathbf{0.0718}\pm0.0185$ \\
        Log-free InCOPE & $0.812\pm0.034$ & $1.134\pm0.085$ & $1.054\pm0.056$ & $0.0934\pm0.0491$ \\
        \midrule
        \multicolumn{5}{@{}l}{\textit{(b) Regression with complete task inputs}} \\
        Full InCOPE & $\mathbf{0.755}\pm0.040$ & $\mathbf{0.659}\pm0.187$ & $\mathbf{0.316}\pm0.073$ & $\mathbf{0.0692}\pm0.0175$ \\
        Ridge & $1.900\pm0.268$ & $2.676\pm1.097$ & $0.517\pm0.026$ & $0.1852\pm0.0373$ \\
        kNN & $1.220\pm0.043$ & $2.379\pm0.086$ & $1.281\pm0.071$ & $0.0777\pm0.0118$ \\
        RBF-SVR & $0.918\pm0.005$ & $4.624\pm0.395$ & $0.410\pm0.022$ & $0.1157\pm0.0057$ \\
        ExtraTrees & $1.036\pm0.050$ & $1.720\pm0.220$ & $0.750\pm0.034$ & $0.0917\pm0.0188$ \\
        \bottomrule
    \end{tabular}
    \vskip -0.2in
\end{table}

\paragraph{Target-policy-only prediction.}
As a complementary control, we test whether target-policy-only prediction can match InCOPE's accuracy. We train ridge regression, distance-weighted kNN, RBF-SVR, and ExtraTrees on source-task pairs of target-policy probabilities and exact policy values. Hyperparameters are selected on source-validation data. Training and testing exclude current-task data and other task inputs. The mean source-training value is a constant reference.

\begin{figure}[htbp]
    \vskip -0.1in
    \centering
    \includegraphics[width=0.9\linewidth]{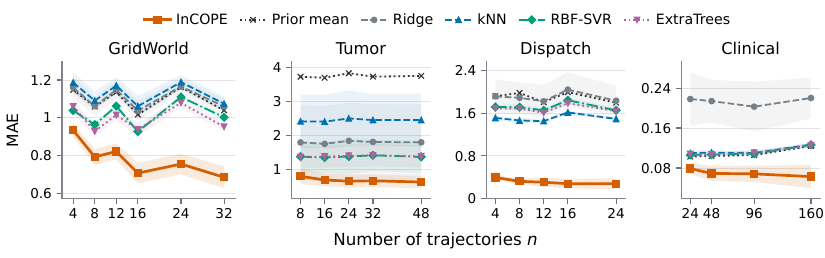}
    \vskip -0.1in
    \caption{Comparison of InCOPE with target-policy-only predictors across trajectory counts.}
    \label{fig:incope-policy-only-log-size}
    \vskip -0.1in
\end{figure}

We group test queries by trajectory count $n$, evaluating all methods on the same queries within each group (Figure~\ref{fig:incope-policy-only-log-size}). InCOPE uses the corresponding offline data, whereas policy-only predictors receive only the target policy at every $n$. Their MAE can vary across $n$ because the groups contain different test queries, not because these predictors receive additional trajectories. Points show mean MAE, and shading shows sample SD across five source-training replicates; vertical axes use domain-specific linear scales. The grouping and evaluation protocol are detailed in Appendix~\ref{app:log-policy-controls}.

Full InCOPE has lower mean MAE than every policy-only control at every tested trajectory count in all four domains, including the smallest datasets. Thus, learning the association between target-policy probabilities and source-task values is insufficient for these regressors to match InCOPE's accuracy. Together with the current-task observation ablation, this result supports using current-task data alongside knowledge acquired during pretraining.

\paragraph{Regression with complete task inputs.}
The policy-only comparison changes both the predictor and its inputs. To match the information available to InCOPE, we give the four regressors a lossless vectorization of its complete preprocessed inputs. They use the same source-training queries and exact value labels as the corresponding specialist. We fix regressor settings and evaluate all methods on a fresh shared test set. Input construction and sample counts are given in Appendix~\ref{app:log-policy-controls}.

InCOPE achieves the lowest mean MAE in all four domains (Table~\ref{tab:incope-information-estimator-controls}(b)). Its advantage over the tested regressors persists when source supervision and available information are matched, supporting InCOPE's structured estimator for learning policy-value prediction across tasks.

\section{Conclusion}
\label{sec:conclusion}

In this paper, we propose InCOPE, a prior-data fitted network for off-policy evaluation in finite-horizon tabular MDPs. By pretraining on synthetic OPE tasks with exact value labels, InCOPE estimates target-policy values from data collected under a behavior policy without task-specific parameter updates. Experiments in four domains show more accurate OPE than the evaluated baselines with limited data, while ablations support combining pretraining with current-task observations.

% \clearpage
\section*{AI use statement}

We used large language models to assist with literature search, manuscript drafting and editing, translation, method development and experimental design, code implementation, mathematical derivations and proofs, and interpretation of results. Their role in synthetic-data generation was to help implement the simulation code; synthetic samples were drawn from the specified simulation models, not generated directly by language models. The manuscript was reviewed and revised, and the key implementation, mathematical derivations and proofs, and experimental results were manually checked. The authors take responsibility for all claims and artifacts reported in this work.

\section*{Reproducibility statement}

Section~\ref{sec:incope} specifies the estimator and pretraining objective, and Appendix~\ref{app:incope-theory} provides the assumptions and proofs for the theoretical results. Appendix~\ref{app:incope-implementation} details input construction, preprocessing, network architecture, and training settings. Appendix~\ref{app:experimental-details} documents the source-task generators and test environments, reference-value computation, baseline implementations, data budgets, evaluation and aggregation procedures, and the comparison and ablation protocols in Section~\ref{sec:experiments}.

\bibliography{iclr2027_conference}
\bibliographystyle{iclr2027_conference}

\appendix
% \section{Appendix}
\section{Theoretical Details}
\label{app:incope-theory}

\subsection{Squared loss and the effective task prior}
\label{app:incope-effective-prior}

The loss scales $s_F$ are fixed positive constants throughout the population
analysis. Define the normalization constant
$Z:=\sum_{F\in\mathcal F}w_F/s_F^2$. The effective task prior can then be
written as
\[
  \Pi_{\mathrm{eff}}
  =\sum_{F\in\mathcal F}\frac{w_F/s_F^2}{Z}\,\Pi_F.
\]
Within each family, the task prior, behavior-policy data distribution,
context map, and exact policy-value labels remain unchanged.
For an arbitrary measurable predictor $f$, write
$\mathcal L_{\mathrm{pre}}(f)$ for the objective in
Eq.~(\ref{eq:incope-population-objective}) with $f_\theta$ replaced by $f$.
Then
\begin{equation}
  \mathcal L_{\mathrm{pre}}(f)
  =Z\,\mathbb E_{\mathrm{eff}}\!\left[(f(C_n)-J)^2\right].
  \label{eq:incope-effective-risk}
\end{equation}
For a specialist, or when all family scales are equal,
$\Pi_{\mathrm{eff}}=\Pi_{\mathrm{pre}}$. This reweighting accounts for the
training objective without assuming that the network observes the family
index. It concerns the population target for fixed scales and does not
assert a finite-pretraining generalization bound.

\subsection{Proof of Proposition~\ref{prop:incope-bayes}}

\begin{proof}
Conditional Jensen's inequality and the law of iterated expectations give
\[
  \mathbb E_{\mathrm{eff}}\!\left[
    \left(\mathbb E_{\mathrm{eff}}[J\mid C_n]\right)^2
  \right]
  \leq
  \mathbb E_{\mathrm{eff}}\!\left[
    \mathbb E_{\mathrm{eff}}[J^2\mid C_n]
  \right]
  =
  \mathbb E_{\mathrm{eff}}[J^2]
  <\infty.
\]
For any square-integrable predictor $f$, the difference
$f(C_n)-\mathbb E_{\mathrm{eff}}[J\mid C_n]$
is measurable with respect to $C_n$.
Since
$\mathbb E_{\mathrm{eff}}[
  \mathbb E_{\mathrm{eff}}[J\mid C_n]-J
  \mid C_n
]=0$,
the law of iterated expectations yields
\[
\begin{aligned}
&\mathbb E_{\mathrm{eff}}\!\left[
  \left(f(C_n)-\mathbb E_{\mathrm{eff}}[J\mid C_n]\right)
  \left(\mathbb E_{\mathrm{eff}}[J\mid C_n]-J\right)
\right]\\
&\quad=
\mathbb E_{\mathrm{eff}}\!\left[
  \mathbb E_{\mathrm{eff}}\!\left[
    \left(f(C_n)-\mathbb E_{\mathrm{eff}}[J\mid C_n]\right)
    \left(\mathbb E_{\mathrm{eff}}[J\mid C_n]-J\right)
    \,\middle|\, C_n
  \right]
\right]\\
&\quad=
\mathbb E_{\mathrm{eff}}\!\left[
  \left(f(C_n)-\mathbb E_{\mathrm{eff}}[J\mid C_n]\right)
  \mathbb E_{\mathrm{eff}}\!\left[
    \mathbb E_{\mathrm{eff}}[J\mid C_n]-J
    \,\middle|\, C_n
  \right]
\right]\\
&\quad=0.
\end{aligned}
\]
Using the effective-risk identity and the vanishing cross term, we obtain
\[
\begin{aligned}
\mathcal L_{\mathrm{pre}}(f)
&=Z\,\mathbb E_{\mathrm{eff}}\!\left[
  (f(C_n)-J)^2
\right]\\
&=Z\,\mathbb E_{\mathrm{eff}}\!\left[
  \left(f(C_n)-\mathbb E_{\mathrm{eff}}[J\mid C_n]\right)^2
\right]
 +Z\,\mathbb E_{\mathrm{eff}}\!\left[
  \left(\mathbb E_{\mathrm{eff}}[J\mid C_n]-J\right)^2
\right]\\
&=Z\,\mathbb E_{\mathrm{eff}}\!\left[
  \operatorname{Var}_{\mathrm{eff}}(J\mid C_n)
\right]
 +Z\,\mathbb E_{\mathrm{eff}}\!\left[
  \left(f(C_n)-\mathbb E_{\mathrm{eff}}[J\mid C_n]\right)^2
\right],
\end{aligned}
\]
where the last equality follows from the definition of conditional
variance and the law of iterated expectations.
The first term is independent of $f$, and the second is nonnegative,
vanishing exactly when
$f(C_n)=\mathbb E_{\mathrm{eff}}[J\mid C_n]$ almost surely.
Hence the minimizer satisfies
$f_{\mathrm{pre}}^\star(C_n)=\mathbb E_{\mathrm{eff}}[J\mid C_n]$
and is unique up to null sets.
\end{proof}

\subsection{Proof of Corollary~\ref{cor:incope-compatibility}}

Let $\mathbb P_{\mathrm{test}}$ be the joint distribution of $(C_n,J)$
induced by test tasks, behavior-policy data generation, and context
construction, and let $\mathbb E_{\mathrm{test}}$ denote expectation
under this distribution.
Write $\mathbb P_{\mathrm{test},C}$ for its context marginal and
$\mathbb P_{\mathrm{eff},C}$ for the context distribution induced by
$\Pi_{\mathrm{eff}}$ and the corresponding data-generation and
context-construction procedures.

The context condition in Corollary~\ref{cor:incope-compatibility} is
precisely
$\mathbb P_{\mathrm{test},C}\ll\mathbb P_{\mathrm{eff},C}$:
for every measurable set $A$ of contexts,
\[
  \mathbb P_{\mathrm{eff},C}(A)=0
  \quad\Longrightarrow\quad
  \mathbb P_{\mathrm{test},C}(A)=0.
\]

\begin{proof}
Proposition~\ref{prop:incope-bayes} establishes that
\[
  f_{\mathrm{pre}}^\star(C_n)
  =
  \mathbb E_{\mathrm{eff}}[J\mid C_n]
  \qquad
  \mathbb P_{\mathrm{eff},C}\text{-almost surely}.
\]
Since
$\mathbb P_{\mathrm{test},C}\ll\mathbb P_{\mathrm{eff},C}$,
this identity also holds
$\mathbb P_{\mathrm{test},C}$-almost surely.
Combining it with the assumed equality
\[
  \mathbb E_{\mathrm{test}}[J\mid C_n]
  =
  \mathbb E_{\mathrm{eff}}[J\mid C_n]
  \qquad
  \mathbb P_{\mathrm{test},C}\text{-almost surely},
\]
we obtain
\[
  f_{\mathrm{pre}}^\star(C_n)
  =
  \mathbb E_{\mathrm{test}}[J\mid C_n]
  \qquad
  \mathbb P_{\mathrm{test},C}\text{-almost surely}.
\]

The test second moment of this predictor satisfies
\begin{equation*}
  \mathbb E_{\mathrm{test}}\!\left[
    f_{\mathrm{pre}}^\star(C_n)^2
  \right]
  =
  \mathbb E_{\mathrm{test}}\!\left[
    \left(\mathbb E_{\mathrm{test}}[J\mid C_n]\right)^2
  \right]
  \leq
  \mathbb E_{\mathrm{test}}\!\left[
    \mathbb E_{\mathrm{test}}[J^2\mid C_n]
  \right]
  =
  \mathbb E_{\mathrm{test}}[J^2]
  <\infty.
\end{equation*}
The first equality uses the identity established above.
The inequality follows from conditional Jensen's inequality.
The second equality follows from the law of iterated expectations.
The final bound is the assumed finite test second moment, so
$f_{\mathrm{pre}}^\star$ is square-integrable under the test distribution.

Let $f$ be any measurable predictor satisfying
$\mathbb E_{\mathrm{test}}[f(C_n)^2]<\infty$.
Since $f_{\mathrm{pre}}^\star(C_n)$ is measurable with respect to $C_n$,
\[
  \mathbb E_{\mathrm{test}}\!\left[
    f_{\mathrm{pre}}^\star(C_n)-J\mid C_n
  \right]
  =
  f_{\mathrm{pre}}^\star(C_n)
  -
  \mathbb E_{\mathrm{test}}[J\mid C_n]
  =0.
\]
The difference $f(C_n)-f_{\mathrm{pre}}^\star(C_n)$ is also measurable
with respect to $C_n$.
Both this difference and the residual
$f_{\mathrm{pre}}^\star(C_n)-J$ are square-integrable, so their product
is integrable by the Cauchy--Schwarz inequality.
The law of iterated expectations and the measurable-factor property
of conditional expectation therefore give
\[
\begin{aligned}
&\mathbb E_{\mathrm{test}}\!\left[
  (f(C_n)-f_{\mathrm{pre}}^\star(C_n))
  (f_{\mathrm{pre}}^\star(C_n)-J)
\right]\\
&\quad=
\mathbb E_{\mathrm{test}}\!\left[
  \mathbb E_{\mathrm{test}}\!\left[
    (f(C_n)-f_{\mathrm{pre}}^\star(C_n))
    (f_{\mathrm{pre}}^\star(C_n)-J)
    \mid C_n
  \right]
\right]\\
&\quad=
\mathbb E_{\mathrm{test}}\!\left[
  (f(C_n)-f_{\mathrm{pre}}^\star(C_n))
  \mathbb E_{\mathrm{test}}\!\left[
    f_{\mathrm{pre}}^\star(C_n)-J\mid C_n
  \right]
\right]\\
&\quad=0.
\end{aligned}
\]

Expanding the squared residual and using the vanishing cross term yields
\[
\begin{aligned}
  \mathbb E_{\mathrm{test}}\!\left[(f(C_n)-J)^2\right]
  &=
  \mathbb E_{\mathrm{test}}\!\left[
    (f(C_n)-f_{\mathrm{pre}}^\star(C_n))^2
  \right]
  +
  \mathbb E_{\mathrm{test}}\!\left[
    (f_{\mathrm{pre}}^\star(C_n)-J)^2
  \right]\\
  &\quad+
  2\mathbb E_{\mathrm{test}}\!\left[
    (f(C_n)-f_{\mathrm{pre}}^\star(C_n))
    (f_{\mathrm{pre}}^\star(C_n)-J)
  \right]\\
  &=
  \mathbb E_{\mathrm{test}}\!\left[
    (f(C_n)-f_{\mathrm{pre}}^\star(C_n))^2
  \right]
  +
  \mathbb E_{\mathrm{test}}\!\left[
    (f_{\mathrm{pre}}^\star(C_n)-J)^2
  \right].
\end{aligned}
\]
All risks above are finite by square integrability.
Subtracting the risk of $f_{\mathrm{pre}}^\star$ gives
\[
\begin{aligned}
  &\mathbb E_{\mathrm{test}}\!\left[(f(C_n)-J)^2\right]
  -
  \mathbb E_{\mathrm{test}}\!\left[
    (f_{\mathrm{pre}}^\star(C_n)-J)^2
  \right]\\
  &\qquad=
  \mathbb E_{\mathrm{test}}\!\left[
    (f(C_n)-f_{\mathrm{pre}}^\star(C_n))^2
  \right]
  \geq0.
\end{aligned}
\]
Since $f$ was arbitrary, $f_{\mathrm{pre}}^\star$ minimizes
the test squared risk over the stated class of predictors.
\end{proof}

The compatibility condition requires agreement of conditional means,
not of complete conditional value distributions or underlying MDPs.
It is a sufficient condition for the ideal prediction rule, not a
consequence of matching marginal value ranges or an assertion that
a particular experimental prior satisfies it.

\section{Implementation Details}
\label{app:incope-implementation}

\subsection{Input Features and Preprocessing}
\label{app:incope-input-features}

Each state--action pair is represented by a 25-dimensional numerical vector
$\mathbf x_{sa}$. The same preprocessing is used for pretraining and evaluation.
Write $S:=|\mathcal S|$ and $A:=|\mathcal A|$.

\paragraph{Trajectory statistics.}
For $D_n=(\tau_1,\ldots,\tau_n)$, let $S_{it}$, $A_{it}$, and $R_{it}$
denote the state, action, and reward at step $t$ of trajectory $i$.
Define
\begin{equation*}
  \begin{aligned}
    N_{sa}
      &:=\sum_{i=1}^{n}\sum_{t=0}^{H-1}
        \mathbf 1\{S_{it}=s,A_{it}=a\},\\
    N_{sas'}
      &:=\sum_{i=1}^{n}\sum_{t=0}^{H-1}
        \mathbf 1\{S_{it}=s,A_{it}=a,S_{i,t+1}=s'\}.
  \end{aligned}
\end{equation*}
The reward mean and standard deviation are
\begin{equation}
  \begin{aligned}
    \widehat r_{sa}
      &:=\frac{\sum_{i,t}R_{it}\mathbf 1\{S_{it}=s,A_{it}=a\}}
              {\max(N_{sa},1)},\\
    \widehat\sigma_{sa}
      &:=\left[
        \max\left\{
          \frac{\sum_{i,t}R_{it}^{2}\mathbf 1\{S_{it}=s,A_{it}=a\}}
               {\max(N_{sa},1)}
          -\widehat r_{sa}^{\,2},\,0
        \right\}
      \right]^{1/2}.
  \end{aligned}
  \label{eq:incope-reward-statistics}
\end{equation}
Both are zero when $N_{sa}=0$. Behavior probabilities and transitions use
the fixed smoothers
\begin{equation}
  \widehat\pi_b(a\mid s)
    :=\frac{N_{sa}+1/2}{\sum_{a'}N_{sa'}+A/2},
  \qquad
  \widehat P(s'\mid s,a)
    :=\frac{N_{sas'}+1/(2S)}{N_{sa}+1/2}.
  \label{eq:incope-empirical-inputs}
\end{equation}
Thus, an unobserved state has a uniform estimated behavior policy, and an
unobserved state--action pair has a uniform estimated transition distribution.

\paragraph{Empirical target visitation.}
Let $p_t(s)$ denote the state probability at step $t$ under $\pi_e$
in the estimated transition model. Starting from $p_0(s):=\rho_0(s)$,
propagate
\begin{equation*}
  p_{t+1}(s')
    :=\sum_{s,a}p_t(s)\pi_e(a\mid s)\widehat P(s'\mid s,a),
  \qquad
  \widehat w_{sa}
    :=\frac{\sum_{t=0}^{H-1}\gamma^t p_t(s)\pi_e(a\mid s)}
           {\sum_{t=0}^{H-1}\gamma^t}.
\end{equation*}
These weights describe target-policy visitation in the empirical model.
Let $v_{sa}:=\mathbf 1\{N_{sa}>0\}$. The count summary used below is
\begin{equation*}
  N_{\mathrm{eff}}
    :=\left[
       \max\left\{\sum_{s,a}\frac{\widehat w_{sa}}{N_{sa}+1/2},
                   10^{-12}\right\}
      \right]^{-1}.
\end{equation*}

\paragraph{Feature vector.}
For a nonnegative array $z$ and $\epsilon>0$, define
\[
  \mathcal N_\epsilon(z)_{sa}
    :=\frac{z_{sa}}{\max\{\max_{s',a'}z_{s'a'},\epsilon\}},
  \qquad
  \operatorname{clip}(z;l,u):=\min\{u,\max\{l,z\}\}.
\]
All feature coordinates are listed in implementation order in
Table~\ref{tab:incope-input-features}. Task-level scalars are repeated across state--action pairs.
The reward coordinates retain their original units.

\begin{table}[t]
  \centering
  \caption{The 25 state--action input features in $\mathbf x_{sa}$.}
  \label{tab:incope-input-features}
  \small
  \setlength{\tabcolsep}{4pt}
  \renewcommand{\arraystretch}{1.24}
  \begin{tabular}{@{}rl@{}}
    \toprule
    Coordinates & Definition \\
    \midrule
    1 & $\log(1+N_{sa})/\log(1+nH)$ \\
    2--3 & $(\widehat r_{sa},\,\widehat\sigma_{sa})$ \\
    4--5 & $(\pi_e(a\mid s),\,\widehat\pi_b(a\mid s))$ \\
    6 & $\log\!\left[\operatorname{clip}\!\left(
          \frac{\pi_e(a\mid s)}{\max\{\widehat\pi_b(a\mid s),10^{-4}\}};
          10^{-4},25\right)\right]/\log 25$ \\
    7--8 & $(\rho_0(s),\,\gamma)$ \\
    9 & $\mathcal N_{10^{-6}}\!\left(
          (\pi_e(a\mid s)\log(1+N_{sa}))_{sa}\right)_{sa}$ \\
    10 & $S^{-1}\sum_{s',a'}\pi_e(a'\mid s')v_{s'a'}$ \\
    11 & $\widehat w_{sa}$ \\
    12 & $\mathcal N_{10^{-8}}\!\left(
           (\widehat w_{sa}\log(1+N_{sa}))_{sa}\right)_{sa}$ \\
    13 & $\mathcal N_{10^{-8}}\!\left(
           (\widehat w_{sa}(1-v_{sa}))_{sa}\right)_{sa}$ \\
    14 & $\mathcal N_{10^{-8}}\!\left(
           (\widehat w_{sa}/\sqrt{N_{sa}+1})_{sa}\right)_{sa}$ \\
    15 & $-\dfrac{\sum_{s'}\widehat P(s'\mid s,a)
           \log\!\left(\max\{\widehat P(s'\mid s,a),10^{-12}\}\right)}
           {\max\{\log S,10^{-8}\}}$ \\
    16--17 & $\left(\sum_{s',a'}\widehat w_{s'a'}v_{s'a'},\,
                    \sum_{s',a'}\widehat w_{s'a'}(1-v_{s'a'})\right)$ \\
    18 & $\log(1+N_{\mathrm{eff}})/\log(1+nH)$ \\
    19--20 & $\left(\log(1+H)/\log 101,\,
                    \sum_{t=0}^{H-1}\gamma^t/100\right)$ \\
    21--24 & $\mathbf d_a$ \\
    25 & $\delta_A$ \\
    \bottomrule
  \end{tabular}
\end{table}

The indicator $\delta_A$ records whether action attributes are supplied.
Clinical uses five dose bins for each of fluid and vasopressor treatment.
For zero-based action indices $a=0,\ldots,24$,
$u_a:=\lfloor a/5\rfloor/4$ and $v_a:=(a\bmod5)/4$ encode the respective
dose bins. The four attributes are
\[
  \mathbf d_a:=(u_a,v_a,(u_a+v_a)/2,u_av_a),
  \qquad \delta_A=1.
\]
The other three domains use $\mathbf d_a=\mathbf 0$ and $\delta_A=0$.

\paragraph{Return scale and batching.}
The output multiplier is computed from the same trajectory statistics:
\begin{equation}
  c(C_n):=
    \max\left\{
      1,\,
      \left(\sum_{t=0}^{H-1}\gamma^t\right)
      \max_{s,a}\bigl(|\widehat r_{sa}|+\widehat\sigma_{sa}\bigr)
    \right\}.
  \label{eq:incope-input-return-scale}
\end{equation}
Unlike the loss scale $s_F$, $c(C_n)$ requires no source labels.
Feature and transition tensors are zero-padded to the largest state and
action counts in each batch. Boolean masks distinguish original entries
from added padding; original state--action pairs with $N_{sa}=0$ remain
unmasked. A separate availability vector is fixed to $(1,0)$, indicating
known $\rho_0$ and estimated $\pi_b$.

\subsection{Network Architecture}
\label{app:incope-network}

Both specialist and unified models use hidden dimension 48, two
action-attention blocks, and $L_S=2$ propagation blocks, with 91,681
trainable parameters. All affine maps include biases. Different blocks
have independent parameters, shared across states and actions within each
block. Dropout is zero, and no positional embeddings are used.

\paragraph{State--action encoding.}
The feature encoder $E_x$ and availability encoder $E_{\mathrm{aux}}$ are
two-layer MLPs with widths $25\!\to\!48\!\to\!48$ and
$2\!\to\!48\!\to\!48$, respectively, each with GELU between its affine
layers. Using the availability vector from
Appendix~\ref{app:incope-input-features}, the input representations are
\begin{equation*}
  h_{sa}^{\mathrm{in}}
    :=E_x(\mathbf x_{sa})+E_{\mathrm{aux}}((1,0)).
\end{equation*}
Each action- or state-attention block has the form
\begin{equation}
  \begin{aligned}
    Y&:=\operatorname{LN}_1\!\left(
         X+\operatorname{MHA}(X,X,X)\right),\\
    \mathcal B(X)&:=\operatorname{LN}_2\!\left(Y+\operatorname{FF}(Y)\right),
  \end{aligned}
  \label{eq:incope-attention-block}
\end{equation}
where $X$ is a sequence of hidden vectors, $\operatorname{MHA}$ is
four-head self-attention with 12 dimensions per head, and
$\operatorname{FF}$ is a $48\!\to\!96\!\to\!48$ MLP with GELU.
Layer normalization $\operatorname{LN}$ acts on each hidden vector;
its parameters are separate for every occurrence.
For each state, two such blocks attend over its actions:
\begin{equation*}
  (h_{sa}^{(0)})_{a\in\mathcal A}
    :=\operatorname{Attn}_{A,2}\!\left(
       \operatorname{Attn}_{A,1}\!\left(
         (h_{sa}^{\mathrm{in}})_{a\in\mathcal A}\right)\right).
\end{equation*}

\paragraph{Policy-conditioned propagation.}
The state-attention operator $\operatorname{Attn}_{S,\ell}$ in
Eq.~\eqref{eq:incope-transition-message} also follows
Eq.~\eqref{eq:incope-attention-block}, with states as the sequence entries.
Within $\Psi_\ell$, the horizon $H$ determines a learned encoding
and an update gate. The encoder $E_H$ is a $1\!\to\!48\!\to\!48$ MLP with GELU,
giving
\begin{equation*}
  e_H:=E_H\!\left(\frac{\log(1+H)}{\log 101}\right),
  \qquad
  g_\ell(H):=\mathbf 1\{\ell<H\}.
\end{equation*}
The integer $H$ is recovered from feature coordinate 19 by inverting its
logarithmic encoding and rounding. Given the state context
$\widetilde u_s^{(\ell)}$ and successor message $m_{sa}^{(\ell)}$ in
Eq.~\eqref{eq:incope-transition-message}, the update $\Psi_\ell$ is
\begin{equation*}
  \begin{aligned}
    h_{sa}^{(\ell+1)}
      ={}&(1-g_\ell(H))h_{sa}^{(\ell)}\\
         &+g_\ell(H)\operatorname{LN}_{\ell}\!\left(
           h_{sa}^{(\ell)}+W_\ell m_{sa}^{(\ell)}+b_\ell
           +\widetilde u_s^{(\ell)}+e_H\right),
    \qquad \ell=0,1,
  \end{aligned}
\end{equation*}
where $W_\ell\in\mathbb R^{48\times48}$ and $b_\ell\in\mathbb R^{48}$
form a learned affine projection of the successor message.
Each propagation block is applied once; blocks with $\ell\ge H$ leave
the representations unchanged. These are hidden-vector updates, not
$H$ steps of scalar Bellman recursion.

\paragraph{Value estimation and masking.}
The prediction head $g_\theta$ in Eq.~\eqref{eq:incope-readout} is a
$48\!\to\!48\!\to\!1$ MLP with GELU and a linear output.
The multiplier $c(C_n)$ is defined in
Eq.~\eqref{eq:incope-input-return-scale}.
Padded entries are excluded as attention keys and values, and their
representations are zeroed after input encoding and every attention or
propagation block. For an entirely padded action sequence, one dummy key
is temporarily unmasked to avoid an undefined softmax; the sequence
output is then zeroed. The target-policy probabilities from feature
coordinate 4 are masked and renormalized within each state. The initial
probabilities from coordinate 7 are averaged over unpadded actions,
masked, and renormalized over states. Normalization denominators for
these probabilities are bounded below by $10^{-8}$.

\subsection{Training Details}
\label{app:incope-training}

\paragraph{Source data.}
For the main comparison, each of the five training replicates uses 3,000
training queries and 600 validation queries per family, generated before
optimization and held fixed throughout the run. Each query pairs an input context with its exact target-policy value;
these counts refer to queries rather than distinct MDPs. Source MDP groups are disjoint between training and
validation. Within each replicate, the specialist and unified models
use the same source queries for each family. Source generation, model
initialization, and training-data order use separate random streams
across the five replicates.

\paragraph{Training objective and scales.}
Training minimizes the objective in Eq.~(\ref{eq:incope-empirical-objective}).
For the main comparison, $M_F=3{,}000$, with $w_F=1$ for the specialist's
single family and $w_F=1/4$ for each family in the unified model.
Writing $\widetilde J_F:=\operatorname{median}_j J_{Fj}$, the default
scale is
\begin{equation*}
  s_F:=1.4826\,\operatorname{median}_j
        |J_{Fj}-\widetilde J_F|.
\end{equation*}
If this value is nonfinite or below $10^{-6}$, the implementation tries
the label interquartile range divided by $1.349$, the label population
standard deviation, and $\max\{|\widetilde J_F|,1\}$ in that order,
accepting the first finite value at least $10^{-6}$.
For a normal distribution, the median absolute deviation (MAD) and
interquartile range are approximately $0.67449$ and $1.34898$ times its
standard deviation, respectively, hence the multiplier $1.4826$ and
divisor $1.349$. This calibration does not require normally distributed
policy-value labels. The threshold $10^{-6}$ and the final fallback
are numerical safeguards against zero or near-zero scales.
The scales are computed separately for each replicate from training
labels only and remain fixed during optimization and validation.
All family scales in the main comparison were computed using the
default MAD rule; no fallback was triggered.

\paragraph{Optimization and model selection.}
Training uses AdamW with peak learning rate $3\times10^{-4}$, weight
decay $10^{-3}$, and gradient-norm clipping at 2.
Four microbatches of eight queries are accumulated before each update,
giving an effective batch size of 32. Each unified-model update contains
one microbatch from each family; a specialist draws all four from its
own family. Queries are shuffled within each family and reshuffled when
the available queries are exhausted. An epoch contains 375 updates
(12,000 draws) for the unified model and 94 updates (3,008 draws) for a
specialist. The learning rate uses one epoch of linear warmup followed
by cosine decay toward $3\times10^{-5}$ over a maximum of 48 epochs.

After every epoch, validation averages the scaled squared errors
over 600 queries per family, using the same family weights and
training-derived scales as Eq.~(\ref{eq:incope-empirical-objective}).
Early stopping is enabled from epoch 8, with patience 8: the patience
counter resets only when validation loss decreases by more than 0.003
relative to its last reset value. The checkpoint with the lowest
validation loss is retained, including improvements smaller than this
threshold. These models are trained from random initialization in PyTorch
on Apple M4 hardware using the MPS backend.

\section{Experimental Details}
\label{app:experimental-details}

\subsection{Task Priors and Test Environments}
\label{app:task-priors}

For each domain, we specify the source MDP generator, the behavior and target policies constructed within an MDP, and the resulting logs and value labels. We then describe the separate test generator. These are the fixed generators used in the reported experiments; source-prior development used separate development evaluations, while checkpoint selection uses source validation as described in Appendix~\ref{app:incope-training}. Source MDP parameters are available to the simulator for generating logs and exact value labels, but are not supplied to the predictor in place of its observed context.

\subsubsection{GridWorld}
\label{app:gridworld}

A GridWorld task is generated by specifying a navigation MDP, constructing
behavior and target policies within that MDP, and sampling behavior logs.
Source and test tasks use the respective environment and policy laws below.

\paragraph{Source environments.}
States represent accessible cells, and actions are north, east, south,
and west. Grid shapes are sampled uniformly from
$\{3\times5,4\times5,5\times5,5\times6,6\times6,6\times7\}$.
The source prior contains four equally represented mechanisms: absorbing
goals and traps, hazard-triggered resets, repeatable rewards, and obstacle
fields. Except in the obstacle mechanism, the graph initially contains
all orthogonally adjacent edges. Edges are considered for removal in
random order toward a fraction sampled from $\mathrm U(0,0.08)$, accepting
only removals that preserve connectivity.
The start $s_0$ is sampled uniformly among cells of degree at least two.
The goal $s_g$ is sampled uniformly among cells at graph distance at least
$\max\{2,\lfloor Q_{0.75}(d(s_0,\cdot))\rfloor\}$ from $s_0$,
where $Q_{0.75}$ denotes the empirical 75th percentile.

Let $T_a(s)$ be the neighboring cell reached along an open edge in
direction $a$, or $s$ if that move is blocked. Before mechanism-specific
changes, the transition law is
\begin{equation}
  P_0(s'\mid s,a)
  =(1-\epsilon-\zeta)\mathbf1\{s'=T_a(s)\}
  +\zeta\mathbf1\{s'=s\}
  +\frac{\epsilon}{3}\sum_{b\ne a}\mathbf1\{s'=T_b(s)\},
  \label{eq:grid-source-transition}
\end{equation}
where $\epsilon\sim\mathrm U(0.03,0.19)$ is the slip probability and
$\zeta\sim\mathrm U(0.01,0.09)$ is the stay probability; renewable-reward
tasks instead use $\zeta\sim\mathrm U(0.04,0.12)$.
The step-cost magnitude is $c\sim\mathrm U(0.003,0.035)$ and the goal
bonus is $b_g\sim\mathrm U(0.48,1.02)$.
Write $P(B\mid s,a)=\sum_{s'\in B}P(s'\mid s,a)$ for a cell set $B$.
Unless stated otherwise, integer counts are uniform over the specified
range, and cell sets are sampled uniformly without replacement.

\begin{enumerate}
  \item \textbf{Episodic exit.}
  Reaching the goal or a trap makes the state absorbing. One to three
  traps are sampled outside the start and goal. Nonabsorbing transitions
  follow $P_0$, and their reward means are
  \[
    \bar r(s,a)=-c+b_gP(s_g\mid s,a)-b_-P(B\mid s,a),
  \]
  where $B$ is the trap set and $b_-\sim\mathrm U(0.35,0.95)$.

  \item \textbf{Reset barrier.}
  Attempted entry into a hazard can send the agent to a reset destination.
  A midpoint of the interior of a shortest start--goal path is designated
  a hazard. One to three candidate hazards are sampled outside $s_0,s_g$
  from cells whose distance from $s_0$ differs by at most one from that
  midpoint's distance. The final hazard set $B$ also includes the midpoint.
  Reset destinations include $s_0$ and, with probability $0.7$, one
  available nonhazard, nongoal cell within distance two of $s_0$.
  Their weights follow $\operatorname{Dirichlet}(\mathbf1)$.
  A fraction sampled from $\mathrm U(0.58,0.94)$ of the $P_0$ mass entering
  hazards is redirected to these destinations. The reward mean is
  \[
    \bar r(s,a)=-c+b_gP(s_g\mid s,a)-b_-P_0(B\mid s,a),
    \qquad b_-\sim\mathrm U(0.55,1.10).
  \]
  The penalty therefore applies before redirection. The goal is absorbing
  with probability $1/2$.

  \item \textbf{Renewable reward.}
  Service cells provide rewards repeatedly without absorption. Their
  number is one, two, or three with probabilities $0.6,0.3,0.1$.
  The first service cell is $s_g$. Each additional one is selected from
  the remaining cells, excluding $s_0$, whose distance to the nearest
  existing service cell is at or above the empirical 70th percentile.
  At service cells, the transition row becomes
  $(1-\delta)P_0+\delta\,\mathbf1\{s'=s\}$, with
  $\delta\sim\mathrm U(0.28,0.56)$; elsewhere it remains $P_0$.
  For service set $B$, let $d(s,B)$ be distance to the nearest service
  cell and define $\Phi(s)=-d(s,B)/\max\{1,\max_{s'}d(s',B)\}$.
  The reward mean is
  \[
    \bar r(s,a)=-c+b_rP(B\mid s,a)
    +\kappa\!\left(\sum_{s'}P(s'\mid s,a)\Phi(s')-\Phi(s)\right),
  \]
  where $b_r\sim\mathrm U(0.38,0.88)$ and
  $\kappa\sim\mathrm U(0.035,0.13)$.

  \item \textbf{Obstacle field.}
  Blocked cells are removed from the state space, leaving an irregular
  connected navigation graph. On an $R\times C$ grid, a local random walk
  proposes up to $20RC$ cell removals toward a count
  $\max\{2,\operatorname{round}(\beta RC)\}$, with
  $\beta\sim\mathrm U(0.12,0.24)$. A removal is accepted only if
  connectivity and at least 12 free cells remain.
  Open edges consist of a random-edge-order spanning tree plus a fraction
  sampled from $\mathrm U(0.18,0.45)$ of the remaining adjacent edges.
  One to three traps outside $s_0,s_g$ are absorbing, and the goal is
  independently absorbing with probability $1/2$.
  Nonabsorbing rows use $P_0$ and the episodic-exit reward formula,
  with $b_-\sim\mathrm U(0.40,1.00)$.
\end{enumerate}

In these source environments, absorbing rows have self-transitions and
zero reward means. The initial distribution assigns $1-\alpha$ to $s_0$
and distributes $\alpha$ among its open-edge neighbors with
Dirichlet$(\mathbf1)$ weights, where $\alpha\sim\mathrm U(0,0.28)$.

\paragraph{Source policies.}
Behavior and target policies are generated within each sampled MDP
without changing its transition or reward law. Each policy draw has its
own action-preference scores $z_{sa}$, constructed from a separately
sampled reference policy as specified below. The scores define the
lower-bounded softmax policy
\begin{equation}
  \pi_{z,T}(a\mid s)
  =\varepsilon_0+(1-4\varepsilon_0)
  \frac{\exp(z_{sa}/T)}{\sum_b\exp(z_{sb}/T)},
  \qquad \varepsilon_0=0.010001.
  \label{eq:grid-source-policy}
\end{equation}
Using $z$ favors higher-scoring actions; using $-z$ reverses this
preference before softmax. Target policies use $z$, whereas behavior
policies use the directions in Table~\ref{tab:grid-source-policies}.
Behavior and target scores are generated independently conditional on
the MDP, rather than obtained by negating one another.

For each policy draw, the reference policy follows
Eq.~\eqref{eq:grid-source-policy} with logits $b_a+e_{sa}$,
$b_a\sim\mathcal N(0,0.28^2)$, $e_{sa}\sim\mathcal N(0,0.18^2)$,
and temperature one. Finite-horizon dynamic programming gives its $Q_0$
values: the expected return from taking each action at time zero and
then following that reference policy. These values are centered within
each state and divided by the median within-state standard deviation
exceeding $10^{-10}$ over action-relevant states, defined below.
The divisor defaults to one if none exists and is floored at $10^{-8}$.
Independent state--action noise $\mathcal N(0,\sigma^2)$, with
$\sigma\sim\mathrm U(0.04,0.12)$, and shared action noise
$\mathcal N(0,0.05^2)$ are added and centered within each state.
The resulting row $\widetilde z_s$ is bounded while preserving its
action ordering:
\begin{equation*}
  z_{sa}=\frac{\widetilde z_{sa}}
  {\max\{1,\max_b|\widetilde z_{sb}|/8\}}.
\end{equation*}
Scores at action-irrelevant states are set to zero.

The temperature controls how concentrated the action probabilities are.
Let $u(a\mid s)=1/4$, $\sigma_s^u=\operatorname{SD}_a Q_0^u(s,a)$,
and $\mu_s^u$ be normalized discounted state visitation under $u$ over
the task horizon. Action-relevant states and their weights are
\[
  \begin{gathered}
  I=\left\{s:\sigma_s^u>
    \max\!\left(10^{-10},10^{-7}
      \max\{\operatorname{median}_{\sigma_{s'}^u>10^{-10}}
        \sigma_{s'}^u,10^{-8}\}\right)\right\},\\
  \nu_s\propto\mu_s^u\sigma_s^u\mathbf1\{s\in I\}.
  \end{gathered}
\]
The weights sum to one; if their unnormalized sum is at most $10^{-14}$,
the factor $\mu_s^u$ is omitted. Action-probability concentration is
measured by $\sum_s\nu_s\max_a\pi_{z,T}(a\mid s)$.
After sampling the desired concentration, $T$ is selected by 72
geometric bisection steps on $[10^{-5},10^4]$ to attain it.
The three target policies use concentrations drawn uniformly from
$[0.36,0.44]$, $[0.48,0.58]$, and $[0.62,0.72]$, respectively.
Behavior-policy concentrations follow Table~\ref{tab:grid-source-policies}.

\begin{table}[t]
  \centering
  \caption{GridWorld source behavior-policy generation.}
  \label{tab:grid-source-policies}
  \small
  \begin{tabular}{@{}lccc@{}}
    \toprule
    Regime & Proportion & Action preference & Concentration \\
    \midrule
    Moderate & $0.22$ & Higher scores ($z$) & $\mathrm U(0.30,0.42)$ \\
    Severe & $0.43$ & Lower scores ($-z$) & $\mathrm U(0.30,0.44)$ \\
    Extreme & $0.35$ & Lower scores ($-z$) & $\mathrm U(0.54,0.70)$ \\
    \bottomrule
  \end{tabular}
\end{table}

\paragraph{Source logs and labels.}
Each sampled MDP and behavior policy produce one log $D_n$ of $n$
trajectories, shared by the three target policies. Rewards are sampled as
\begin{equation}
  R_t=\bar r(S_t,A_t)+\varepsilon_t,
  \qquad \varepsilon_t\sim\mathcal N(0,0.03^2),
  \label{eq:grid-observed-reward}
\end{equation}
independently of the successor conditional on $(S_t,A_t)$.
Every trajectory contains $H$ transitions, including after absorption,
and $\gamma=0.95$. Each target policy receives its own exact
finite-horizon value label $J(\mathcal M,\pi_e)$ from
Eq.~\eqref{eq:ope-estimand}, computed by dynamic programming in the source MDP.
Each replicate has 1,000 training and 200 validation MDP/log groups,
giving the 3,000/600 queries in Appendix~\ref{app:incope-training}.
Within each split, the four mechanisms occur equally often;
$H\in\{12,18,24,32\}$ and $n\in\{4,8,12,16,24,32\}$ have marginal
counts differing by at most one, and the behavior-regime proportions
are exact.

\paragraph{Test environments.}
Testing uses a separate $4\times4$ generator with 16 state indices and
the same four actions. Walls prevent entry but are retained among the
state indices. The four modes and parameter ranges appear in
Table~\ref{tab:grid-test-environments}; parameter intervals and integer
count ranges specify uniform distributions, and cells are sampled
uniformly without replacement from the eligible sets.
Except in cliff tasks, the start is a uniformly sampled corner and
the goal is diagonally opposite.
\begin{enumerate}
  \item \textbf{Sparse.} One to four traps are placed outside the start
  and goal. The goal is absorbing with probability $0.55$, and all traps
  are made absorbing together with probability $0.35$.
  \item \textbf{Cliff.} The start and goal are the lower-left and
  lower-right corners. One or two traps are sampled from the two
  intervening cells. A task-level switch, enabled with probability $0.75$,
  makes every trap entry reset to the start. The goal is absorbing
  with probability $0.35$.
  \item \textbf{Dense.} Zero to three traps are placed outside the start
  and goal. All states are nonabsorbing, and movement toward the goal
  contributes an additional reward term specified below.
  \item \textbf{Maze.} One to four walls are sampled outside the start
  and goal, followed by one to four traps from the remaining nonwall
  cells outside the start and goal. The goal is absorbing with
  probability $0.45$.
\end{enumerate}

\begin{table}[t]
  \centering
  \caption{GridWorld test-mode probabilities and uniform parameter ranges.}
  \label{tab:grid-test-environments}
  \small
  \setlength{\tabcolsep}{3pt}
  \begin{tabular}{@{}lcccccc@{}}
    \toprule
    Mode & Probability & Slip $\epsilon$ & Stay $\zeta$ & Cost $c$
      & Bonus $b_g$ & Penalty $b_-$ \\
    \midrule
    Sparse & $.31$ & $[.02,.20]$ & $[0,.04]$ & $[0,.055]$
      & $[.45,1.25]$ & $[.35,1.15]$ \\
    Cliff & $.25$ & $[.02,.16]$ & $[0,.04]$ & $[.015,.065]$
      & $[.70,1.55]$ & $[.70,1.75]$ \\
    Dense & $.20$ & $[0,.18]$ & $[0,.05]$ & $[0,.035]$
      & $[.25,.95]$ & $[.20,.85]$ \\
    Maze & $.24$ & $[.02,.22]$ & $[0,.06]$ & $[0,.05]$
      & $[.40,1.20]$ & $[.35,1.20]$ \\
    \bottomrule
  \end{tabular}
\end{table}

Let $T_a(s)$ denote the boundary- and wall-aware move before any cliff
reset. Unlike Eq.~\eqref{eq:grid-source-transition}, test slip is split
between the two perpendicular directions: the movement probabilities
are $p_a=1-\epsilon-\zeta$ and $p_{a-1}=p_{a+1}=\epsilon/2$, with indices
modulo four, plus stay probability $\zeta$.
When cliff reset is enabled, a movement into a trap leads to $s_0$
instead. Absorbing states have self-transitions.
For a nonabsorbing state, the pre-perturbation reward mean is
\begin{equation*}
  \begin{aligned}
    r_0(s,a)=-c+\sum_{b\in\{a,a-1,a+1\}}p_b
    \bigl[&-b_-\mathbf1\{T_b(s)\in B\}
      +b_g\mathbf1\{T_b(s)=s_g\}\\
      &+0.05\bigl(D(s)-D(T_b(s))\bigr)
      \mathbf1\{\text{dense},\ T_b(s)\notin B\cup\{s_g\}\}\bigr],
  \end{aligned}
\end{equation*}
where $B$ is the trap set and $D$ is Manhattan distance to the goal.
Absorbing rows initially have $r_0=0$. With probability $0.35$, independent
$\mathrm U(-0.015,0.015)$ perturbations are added to all entries,
including absorbing rows. Writing the perturbation as $\xi$ (zero
otherwise), the final mean is
$\bar r=(r_0+\xi)/\max\{1,\max_{s,a}|r_0(s,a)+\xi(s,a)|\}$.
The initial state is $s_0$, except with probability $0.15$, when it is
uniform over the distinct states in $\{s_0,T_0(s_0),\ldots,T_3(s_0)\}$.

\paragraph{Test policies.}
Test policies are constructed from $Q_0^u$, the time-zero action value
under the uniform policy $u(a\mid s)=1/4$, computed by finite-horizon
dynamic programming. Define the statewise softmax policies
\[
  \pi_c=\operatorname{softmax}(Q_0^u/T_c),\quad
  \pi_w=\operatorname{softmax}(Q_0^u/T_w),\quad
  \pi_-=\operatorname{softmax}(-Q_0^u/T_-),
\]
with $T_c\sim\mathrm U(0.15,0.5)$, $T_w\sim\mathrm U(0.7,1.5)$,
and $T_-\sim\mathrm U(0.2,0.8)$.
The first two favor larger $Q_0^u$ values with different temperature
ranges; the third favors smaller values. Their mixtures with $u$ in
Table~\ref{tab:grid-test-policies} define the stationary policy pairs.

\begin{table}[t]
  \centering
  \caption{GridWorld test-policy mixtures.}
  \label{tab:grid-test-policies}
  \small
  \begin{tabular}{@{}lccc@{}}
    \toprule
    Regime & Probability & Target policy & Behavior policy \\
    \midrule
    Moderate & $.22$ & $.90\pi_c+.10u$ & $.55\pi_w+.45u$ \\
    Severe & $.43$ & $.95\pi_c+.05u$ & $.65\pi_-+.35u$ \\
    Extreme & $.35$ & $.98\pi_c+.02u$ & $.90\pi_-+.10u$ \\
    \bottomrule
  \end{tabular}
\end{table}

\paragraph{Test logs and reference values.}
For each test MDP and policy pair, $D_n$ is sampled under $\pi_b$ and
the reference value $J(\mathcal M,\pi_e)$ is computed by exact
finite-horizon dynamic programming. Test horizons and trajectory counts are
sampled uniformly from the source sets. Reward sampling follows
Eq.~\eqref{eq:grid-observed-reward}, with $\gamma=0.95$ and $H$ transitions
per trajectory, including after absorption.
The main comparison comprises five blocks of 500 separately generated
MDP/policy/log queries, totaling 2,500, without shared-log query groups.

\subsubsection{Tumor}
\label{app:tumor-environments}

Tumor tasks model treatment decisions. The four actions $(a_c,a_r)$ are
$(0,0)$, $(1,0)$, $(0,1)$, and $(1,1)$, indicating whether chemotherapy
and radiotherapy are administered. Both source and test tasks
use $H=25$ and $\gamma=0.97$, but differ in their state representation and
growth and treatment mechanisms.

\paragraph{Source environments.}
The state is a pair $(x,z)$ of tumor burden and accumulated toxicity.
Burden is discretized into $B\in\{8,12,16\}$ equal-width bins on $[0,1.25]$,
and toxicity into three equal-width bins on $[0,1]$. Bin midpoints represent
the states, giving $3B\in\{24,36,48\}$ states. The untreated growth
increment is one of
\begin{equation*}
 g(x)=\begin{cases}
   r_gx\log(K/\max\{x,10^{-8}\}),&\text{Gompertz},\\
   \min\{r_ex,\ell\},&\text{exponential--linear},\\
   \alpha_g x^p-\beta_gx,&\text{generalized Bertalanffy}.
 \end{cases}
\end{equation*}
The source design balances these three laws and the three burden
resolutions. Table~\ref{tab:tumor-source-parameters} specifies the parameter ranges. Each parameter is fixed within an MDP. Continuous parameters use independently permuted stratified
midpoints in their respective ranges, rather than independent draws for
each task. Within each growth-law and allocation stratum, resolution-wise
rank grids are interleaved into one grid; training and validation use
separate allocations.

The chemotherapy and radiotherapy kill fractions are
\begin{equation*}
 k_c(x)=\frac{E_{\max}d_c}{E_{50}+d_c}
 \left(1-\omega+\omega\frac{[g(x)]_+}
 {\max\{10^{-12},\max_j[g(x_j)]_+\}}\right),\qquad
 k_r=1-\exp(-\alpha_rd_r-\beta_rd_r^2),
\end{equation*}
where $x_j$ ranges over burden-bin midpoints and $[v]_+=\max\{v,0\}$.
For each state--action pair, 96 independent pairs of Gaussian noise draws
produce
\begin{align*}
 x'&=\operatorname{clip}_{[0,1.25]}
 \left(x+g(x)-x\{1-(1-a_ck_c(x))(1-a_rk_r)\}+\epsilon_x\right),\\
 z'&=\operatorname{clip}_{[0,1]}
 \left((1-\delta)z+a_c\tau_c+a_r\tau_r+\epsilon_z\right),
\end{align*}
with $\epsilon_x\sim\mathcal N(0,\sigma_x^2)$ and
$\epsilon_z\sim\mathcal N(0,\sigma_z^2)$. Successor-bin frequencies define
$P$. The unnormalized reward mean is the average over the same 96 draws of
\begin{equation*}
 -x'/1.25-w_z z'-c_ca_c-c_ra_r+0.15\mathbf1\{x'<0.08\}.
\end{equation*}
The resulting reward matrix is divided by its maximum absolute entry
(with denominator floor $10^{-12}$). The initial distribution is the
product of burden weights proportional to $u_j^4(1-u_j)$, where
$u_j=x_j/1.25$, and toxicity probabilities $(0.85,0.15,0)$.

\begin{table}[t]
\centering
\caption{Tumor source parameter ranges.}
\label{tab:tumor-source-parameters}
\small
\setlength{\tabcolsep}{5pt}
\begin{tabular}{@{}llll@{}}
\toprule
Parameter & Range & Parameter & Range \\
\midrule
Gompertz $r_g$ & $[0.025,0.090]$ & Capacity $K$ & $[0.90,1.40]$ \\
Exponential rate $r_e$ & $[0.04,0.12]$ & Linear cap $\ell$ & $[0.015,0.060]$ \\
Anabolic rate $\alpha_g$ & $[0.04,0.12]$ & Catabolic rate $\beta_g$ & $[0.02,0.09]$ \\
Growth power $p$ & $[0.55,0.90]$ & Chemotherapy dose $d_c$ & $[0.35,1.00]$ \\
Maximum effect $E_{\max}$ & $[0.04,0.18]$ & Half-effect dose $E_{50}$ & $[0.15,0.75]$ \\
Growth-response blend $\omega$ & $[0.35,0.85]$ & Radiotherapy dose $d_r$ & $[0.50,2.50]$ \\
Linear radiation effect $\alpha_r$ & $[0.03,0.16]$ & Quadratic effect $\beta_r$ & $[0.005,0.045]$ \\
Toxicity recovery $\delta$ & $[0.10,0.35]$ & Chemotherapy toxicity $\tau_c$ & $[0.08,0.25]$ \\
Radiotherapy toxicity $\tau_r$ & $[0.04,0.18]$ & Burden noise SD $\sigma_x$ & $[0.005,0.030]$ \\
Toxicity noise SD $\sigma_z$ & $[0.005,0.025]$ & Toxicity weight $w_z$ & $[0.10,0.35]$ \\
Chemotherapy cost $c_c$ & $[0.01,0.08]$ & Radiotherapy cost $c_r$ & $[0.01,0.08]$ \\
\bottomrule
\end{tabular}
\end{table}

\paragraph{Source policies and logs.}
Let $U(a\mid s)=1/4$. Finite-horizon evaluation of $U$ in the generated MDP
gives its time-zero action values $Q_0^U$. Define the two stationary policies
\begin{equation}
 \begin{aligned}
 \pi^Q(a\mid s)&=\operatorname{softmax}_a(Q_0^U(s,a)/0.10),\\
 \pi^C(a\mid s)&=\operatorname{softmax}_a(v_a/0.45),\qquad
 v=(1.1,0.35,0.30,-0.25).
 \end{aligned}
 \label{eq:tumor-policy-components}
\end{equation}
Thus, $\pi^Q$ is constructed from uniform-policy action values, while
$\pi^C$ favors less intensive treatment. They are combined with $U$ as in
Table~\ref{tab:tumor-policy-pairs}. The low-, moderate-, and severe-shift
regimes have proportions $0.20,0.45,0.35$; the standard, broader, and sharper
mixtures have proportions $0.50,0.25,0.25$. The finite allocation preserves
both margins, rounding the joint counts when necessary. Policies are
clipped below at $10^{-8}$ and normalized within each state.

\begin{table}[t]
\centering
\caption{Behavior--target policy pairs for Tumor tasks.}
\label{tab:tumor-policy-pairs}
\small
\setlength{\tabcolsep}{6pt}
\begin{tabular}{@{}llll@{}}
\toprule
Mixture & Regime & Target policy $\pi_e$ & Behavior policy $\pi_b$ \\
\midrule
Standard & Low & $0.75\pi^Q+0.25\pi^C$ & $0.75\pi_e+0.25U$ \\
& Moderate & $0.90\pi^Q+0.10U$ & $0.55\pi^C+0.45U$ \\
& Severe & $0.96\pi^Q+0.04U$ & $0.80\pi^C+0.20U$ \\
Broader & Low & $0.65\pi^Q+0.35\pi^C$ & $0.60\pi_e+0.40U$ \\
& Moderate & $0.82\pi^Q+0.18U$ & $0.40\pi^C+0.60U$ \\
& Severe & $0.90\pi^Q+0.10U$ & $0.65\pi^C+0.35U$ \\
Sharper & Low & $0.85\pi^Q+0.15\pi^C$ & $0.90\pi_e+0.10U$ \\
& Moderate & $0.97\pi^Q+0.03U$ & $0.70\pi^C+0.30U$ \\
& Severe & $0.99\pi^Q+0.01U$ & $0.93\pi^C+0.07U$ \\
\bottomrule
\end{tabular}
\end{table}

Each training replicate contains 500 source MDPs and 100 validation MDPs.
Each MDP is assigned two distinct trajectory counts from
$\{8,16,24,32,48\}$, with the ten unordered pairs represented equally.
For each trajectory count, three policy pairs are assigned and a fresh behavior log
is generated for each pair. This produces 3,000 training and 600 validation
queries.
The environment allocation balances the three growth laws, the three
resolutions, and an auxiliary three-level parameter-stratification index;
the 27 cells contain 18 or 19 training MDPs and three or four validation MDPs.
The policy-regime allocation is separate from this environment allocation.
Every trajectory has 25 steps, with observed rewards drawn independently
as $R_t\sim\mathcal N(\bar r(S_t,A_t),0.02^2)$ after the tabular reward
matrix has been fixed. Each label is the finite-horizon value computed by
dynamic programming in that generated tabular MDP.

\paragraph{Test environments, policies, and logs.}
The test state is tumor burden alone, with 24 representative values
$x_s=0.02+0.98s/23$, $s=0,\ldots,23$. Successors are assigned to 24
equal-width bins on $[0,1.05]$. For each state--action pair, 96 draws use
\begin{equation*}
 x'=\operatorname{clip}_{[0,1.05]}
 \left[x+r_tx(1-x/K_t)-\eta_ca_cx-\eta_ra_r\sqrt{x}
       -\eta_{cr}a_ca_r+\epsilon\right],
 \qquad \epsilon\sim\mathcal N(0,\sigma_t^2).
\end{equation*}
The seven task-level parameters are drawn independently as
\[
\begin{aligned}
 r_t&\sim\mathrm U(0.045,0.105),& K_t&\sim\mathrm U(0.88,1.18),\\
 \eta_c&\sim\mathrm U(0.035,0.090),& \eta_r&\sim\mathrm U(0.025,0.075),\\
 \eta_{cr}&\sim\mathrm U(0.004,0.018),& \sigma_t&\sim\mathrm U(0.015,0.040),\\
 w_t&\sim\mathrm U(0.03,0.08).&&
\end{aligned}
\]
Successor frequencies define $P$, and $\bar r(s,a)$ is the 96-draw average
of $-x'-w_t(a_c+a_r)+0.15\mathbf1\{x'<0.08\}$, without the source reward
normalization. Four distinct initial states are sampled uniformly from
$\{8,\ldots,15\}$ and assigned $\operatorname{Dirichlet}(\mathbf1_4)$
probabilities. The policies in Eq.~\eqref{eq:tumor-policy-components} are
recomputed in each test MDP, using only the standard mixture in
Table~\ref{tab:tumor-policy-pairs}. Regime probabilities remain
$(0.20,0.45,0.35)$, and $n$ is uniform on $\{8,16,24,32,48\}$.
Behavior trajectories use the same $H$, $\gamma$, and reward-noise SD as
source tasks. Evaluation labels are exact values of the resulting tabular
MDP, not Monte Carlo estimates from the behavior trajectories.

\subsubsection{Dispatch}
\label{app:dispatch-environments}

Dispatch models a two-sided vehicle-service market. A state is a pair
$s=(z,j)$ of a market zone and an ordered supply--demand imbalance level.
The four actions are $(u_a,v_a)\in\{(0,0),(1,0),(0,1),(1,1)\}$, where
$u_a$ enables a driver-supply incentive and $v_a$ enables a passenger-demand
promotion. Source and test environments share these semantics but use
different zone dynamics and imbalance kernels.

\paragraph{Source environments.}
The zone/imbalance-level counts $(Z,B)$ are $(6,4)$, $(7,3)$, $(8,3)$,
$(8,4)$, or $(9,3)$. Write
$x_j=-1+2j/(B-1)$, $j=0,\ldots,B-1$, for source imbalance levels.
A uniformly permuted cycle connects the zones. The graph is either this
cycle, a circulant graph adding all edges at one uniformly chosen cycle
offset in $\{2,\ldots,\lfloor Z/2\rfloor\}$, or the cycle augmented with
$\lfloor Z/3\rfloor$ uniformly sampled noncycle edges. Each zone has
self-transition probability $c_z\sim\mathrm U(0.42,0.76)$; its remaining
mass is distributed over graph neighbors using independent
$\operatorname{Gamma}(1.8,1)$ weights normalized to sum to one.
Denote the resulting zone kernel by $K_{zz'}$ and graph distance by
$d_G(z,z')$.

Demand $d_z$ and supply $b_z$ are independent positive graph fields.
For each field, draw $M$ uniformly from $\{1,2,3\}$, choose $M$ distinct
centers $c_m$ uniformly, and set
\begin{equation*}
  \begin{aligned}
  f_z&=0.20+g_z+\sum_{m=1}^{M}A_m\exp\{-d_G(z,c_m)/w_m\},\\
  g_z&\sim\operatorname{Gamma}(1.7,0.12),\quad
  A_m\sim\mathrm U(0.45,1.45),\quad w_m\sim\mathrm U(0.55,1.85).
  \end{aligned}
\end{equation*}
Each field is divided by its mean over zones. Let
$h_z=0.30+0.70\operatorname{rank}_z/(Z-1)$, where
$\operatorname{rank}_z\in\{0,\ldots,Z-1\}$ orders
$\log(d_z/b_z)$ increasingly, breaking ties by zone index.
The prior contains three equally represented imbalance mechanisms.
Their action-dependent scores are
\begin{equation*}
\begin{aligned}
 q^{\mathrm{M}}_{zj,a}
   &=h_z+p x_j+w\log(d_z/b_z)+0.20v_a-0.38u_a,\\
 q^{\mathrm{Q}}_{zj,a}
   &=h_z+(1-\lambda)p x_j
      +\lambda w\frac{d_z e^{\kappa x_j}-b_z e^{-\kappa x_j}}
                         {d_z e^{\kappa x_j}+b_z e^{-\kappa x_j}}
      +0.20v_a-0.38u_a,\\
 q^{\mathrm{F}}_{zj,a}
   &=h_z+p x_j+w\sum_{z'}K_{zz'}
      \log\frac{d_{z'}e^{\kappa x_j}}
      {(1-\varphi)b_{z'}+\varphi K_{zz'}\sum_k b_k}
      +0.20v_a-0.38u_a.
\end{aligned}
\end{equation*}
Here M, Q, and F denote matching, queue relaxation, and conserved-fleet
coupling, respectively. The last mechanism redistributes the same total
supply toward the zone-transition probabilities. The parameter ranges appear in Table~\ref{tab:dispatch-source-parameters}. Uniform draws are independent except for the conditional range of $c_v$.

The score determines one of three response locations:
\begin{equation*}
  \ell(q)=
  \begin{cases}
    -A, & q<0.45,\\
    \ell_0, & 0.45\le q\le0.78,\\
    A, & q>0.78.
  \end{cases}
  \qquad
  P((z',j')\mid(z,j),a)=K_{zz'}k_m(x_{j'};\ell(q^m_{zj,a}),\sigma),
\end{equation*}
The kernels $k_m$ are normalized over the $B$ levels and are proportional
to $\exp[-(x-\ell)^2/(2\sigma^2)]$ for M,
$\exp[-|x-\ell|/\sigma]$ for Q, and
$[1+(x-\ell)^2/\sigma^2]^{-1.75}$ for F.
Starting from $A_0\sim\mathrm U(0.20,0.25)$, the amplitude $A$ is reduced
only if necessary so that
\[
 \left|\sum_{j'}\frac{x_{j'}}2
    \{k_m(x_{j'};A,\sigma)-k_m(x_{j'};-A,\sigma)\}\right|
 \le 0.219-72\epsilon_{\mathrm{fp32}}.
\]
The largest admissible amplitude in $[|\ell_0|,A_0]$ is found by 64
bisection steps, with the displayed margin protecting the bound after
float32 conversion, where $\epsilon_{\mathrm{fp32}}=2^{-23}$.

For severity $e_j=(|x_j|-\min_k|x_k|)/(1-\min_k|x_k|)$, the raw reward is
\begin{equation*}
\begin{aligned}
 r_{\mathrm{raw}}((z,j),a)
   ={}&s_0-(s_1+c_I)e_j+t_m(2e_j-1)+\delta_z\\
     &-c_u u_a-c_v v_a-c_{uv}u_av_a
       +b_u u_a\max(x_j/2,0)+b_v v_a\max(-x_j/2,0).
\end{aligned}
\end{equation*}
It combines service surplus, imbalance and intervention costs, and
state-dependent benefits of supply or demand interventions. Before
sampling logs, a fixed transformation adjusts the state-dependent
no-intervention baseline without changing the raw action contrasts:
\begin{equation*}
\begin{aligned}
 r_0(s)&=r_{\mathrm{raw}}(s,0),\qquad M_0=\max_s|r_0(s)|,\\
 \widetilde r(s,a)&=\alpha_m r_0(s)+(1-\alpha_m)M_0
                    +r_{\mathrm{raw}}(s,a)-r_0(s),\\
 \bar r(s,a)&=\frac{\widetilde r(s,a)}
                   {\max\{10^{-12},\max_{s',a'}|\widetilde r(s',a')|\}},
 \qquad \alpha_m=\frac{1-0.74}{1-\mu_m}.
\end{aligned}
\end{equation*}
The fixed source-calibration constants are
\[
\begin{aligned}
\mu_{\mathrm M}&=0.7180710724510332,\qquad
\mu_{\mathrm Q}=0.7416735225049728,\\
\mu_{\mathrm F}&=0.7945860496332063.
\end{aligned}
\]
They are mechanism-specific means of
$J(\mathcal M,\pi_e)/\sum_{t=0}^{H-1}\gamma^t$ in a fixed source-only
reference collection used when specifying the reward generator.
This transformation is part of the source reward generator, not a
post-processing of predicted policy values.

\begin{table}[t]
\centering
\caption{Dispatch source parameter distributions.}
\label{tab:dispatch-source-parameters}
\small
\begin{tabular}{@{}lll@{}}
\toprule
Parameter & Meaning & Distribution or value \\
\midrule
$p$ & Imbalance persistence & $\mathrm U(0.16,0.24)$ \\
$w$ & Market-pressure weight & $\mathrm U(0.03,0.08)$ \\
$\kappa$ & State-pressure strength & $\mathrm U(0.12,0.28)$ \\
$\lambda$ & Queue relaxation & $\mathrm U(0.44,0.70)$ \\
$\varphi$ & Fleet reallocation fraction & $\mathrm U(0.24,0.48)$ \\
$A_0$ & Unconstrained response amplitude & $\mathrm U(0.20,0.25)$ \\
$\ell_0$ & Neutral response location & $\mathrm U(0,0.035)$ \\
$\sigma$ & Ordinal-kernel scale & $\mathrm U(0.55,0.82)$ \\
$s_0$ & Service-surplus level & $\mathrm U(0.98,1.04)$ \\
$s_1$ & Service loss at extreme imbalance & $\mathrm U(0.42,0.48)$ \\
$c_I$ & Cost at extreme imbalance & $\mathrm U(0.07,0.11)$ \\
$b_u$ & Driver balance benefit & $\mathrm U(0.22,0.30)$ \\
$b_v$ & Passenger balance benefit & $\mathrm U(0.16,0.24)$ \\
$c_u$ & Driver incentive cost & $\mathrm U(0.09,0.17)$ \\
$c_v$ & Passenger promotion cost
  & $\mathrm U(0.12,\min\{0.23,c_u+b_v/2-0.015\})$ \\
$c_{uv}$ & Joint intervention cost & $\mathrm U(0.005,0.020)$ \\
$t_m$ & Mechanism service adjustment & $(-0.012,0,0.012)$ for (M,Q,F) \\
$\delta_z$ & Zone service adjustment & $\mathrm U(-0.015,0.015)$ \\
\bottomrule
\end{tabular}
\end{table}

\paragraph{Source initial distributions.}
Each initial distribution factors as $\rho_0(z,j)=q_zg_j$. One fifth of
the $(8,3)$ tasks start uniformly over zones at the central imbalance
level. The remaining tasks cycle through three initial-distribution
types within each shape and mechanism: (i) $q\sim
\operatorname{Dirichlet}(c\mathbf1)$ with $c\sim\mathrm U(8,32)$, and
$g$ concentrated at the central level, or split between the two central
levels with lower-level mass $\mathrm U(0.35,0.65)$ when $B=4$;
(ii) $q_z=1/Z$ and $g_j\propto\exp[-x_j^2/(2\omega^2)]$ with
$\omega\sim\mathrm U(0.20,0.55)$; and
(iii) $q_z\propto\sqrt{d_zb_z}\exp(\varepsilon_z)$ with
$\varepsilon_z\sim\mathcal N(0,0.18^2)$, and
$g_j\propto\exp[-(x_j-\ell)^2/(2\omega^2)]$, where
$\ell=\operatorname{clip}(0.35Z^{-1}\sum_z\log(d_z/b_z),-0.65,0.65)$
and $\omega\sim\mathrm U(0.24,0.62)$.

\paragraph{Source policies, logs, and labels.}
Let $Q_0^u$ be the finite-horizon time-zero action values of the uniform
policy $u(a\mid s)=1/4$ in the sampled source MDP. Policies have the form
\begin{equation*}
  \pi_{\pm,T}(a\mid s)
  =0.98\frac{\exp(\pm Q_0^u(s,a)/T)}
                 {\sum_b\exp(\pm Q_0^u(s,b)/T)}+0.005.
\end{equation*}
The target policy favors higher action values, while the behavior policy
favors lower values. For each policy, 64 geometric bisection steps over
$[10^{-4},10^3]$ select a temperature matching its prescribed concentration,
$\sum_s\rho_0(s)\max_a\pi(a\mid s)$.
Target concentrations are $\{0.30,0.38,0.48,0.62\}$; behavior
concentrations are $\{0.27,0.32,0.38,0.46\}$ in the severe regime and
$\{0.29,0.36,0.44,0.52\}$ in the extreme regime. The two regimes are
equally represented, and the behavior and target concentration indices
are balanced separately.

Each replicate contains 3,000 training and 600 validation queries,
obtained from two independent draws of the same 1,500/300-query design.
Within each draw, a full factorial design over the five shapes,
$H\in\{24,32,40\}$, $n\in\{4,8,12,16,24\}$, and the four target
concentrations is repeated five times for training and once for
validation. Topology, behavior regime, and behavior concentration are
balanced, and the three imbalance mechanisms occur equally often
within each shape. Each query uses its own sampled MDP, one behavior
log, and one target policy. The $n$ trajectories each contain $H$
transitions, with rewards sampled independently of the successor as
$R_t\sim\mathcal N(\bar r(S_t,A_t),0.03^2)$ and discount $\gamma=0.97$.
The label is the exact finite-horizon value of that query's target policy,
computed by dynamic programming in its sampled MDP.

\paragraph{Test environments.}
Testing uses eight zones on a fixed cycle and three imbalance levels
$y_j=(j-1)/2\in\{-1/2,0,1/2\}$. For
$\theta\sim\mathrm U(0,2\pi)$ and $\eta\sim\mathrm U(0.18,0.36)$,
the zone demand score and unperturbed zone kernel are
\[
 d_z^{\mathrm{test}}=0.65+0.35\sin(2\pi z/8+\theta),\qquad
 K^0_{zz'}=(1-\eta)\mathbf1\{z'=z\}
   +\frac\eta2\mathbf1\{z'=z-1\text{ or }z+1\},
\]
with zone indices interpreted modulo eight. Independently for each
state--action row, with probability $0.15$ a mass of $0.05$ is added at
a uniformly sampled zone and the row is renormalized; otherwise it
remains $K^0$. Denote this row by $K_{zj,a,z'}^{\mathrm{test}}$.
The imbalance response is
\begin{equation*}
\begin{aligned}
 q_{zj,a}^{\mathrm{test}}
   &=d_z^{\mathrm{test}}+0.40y_j+0.20v_a-0.38u_a,\\
 g(q)&=
 \begin{cases}
  (0.30,0.56,0.14),&q<0.45,\\
  (0.20,0.58,0.22),&0.45\le q\le0.78,\\
  (0.12,0.56,0.32),&q>0.78,
 \end{cases}\\
 P((z',j')\mid(z,j),a)
   &=K_{zj,a,z'}^{\mathrm{test}}g_{j'}(q_{zj,a}^{\mathrm{test}}).
\end{aligned}
\end{equation*}
For independent $c_{sa}\sim\mathrm U(0.08,0.22)$ and
$\xi_{sa}\sim\mathcal N(0,0.02^2)$, sampled once when constructing the
MDP, its raw reward table is
\begin{equation*}
 r_{\mathrm{test}}((z,j),a)
 =1-1.08|y_j|-c_{sa}(u_a+v_a)
  +0.28u_a\max(y_j,0)+0.18v_a\max(-y_j,0)+\xi_{sa}.
\end{equation*}
The reward means equal this table divided by
$\max\{10^{-6},\max_{s,a}|r_{\mathrm{test}}(s,a)|\}$.
The initial distribution is uniform over zones at neutral imbalance:
$\rho_0(z,j)=\mathbf1\{j=1\}/8$.

\paragraph{Test policies, logs, and labels.}
Let $Q_0^u$ again denote time-zero action values of the uniform policy,
now in the test MDP, and define
$\pi_+=\operatorname{softmax}(Q_0^u/T_+)$ and
$\pi_-=\operatorname{softmax}(-Q_0^u/T_-)$, with
$T_+\sim\mathrm U(0.15,0.50)$ and
$T_-\sim\mathrm U(0.20,0.80)$, applied separately within each state.
With probability $0.45$, the severe regime uses
$(\pi_e,\pi_b)=(0.95\pi_++0.05u,\,0.65\pi_-+0.35u)$;
otherwise, the extreme regime uses
$(\pi_e,\pi_b)=(0.98\pi_++0.02u,\,0.90\pi_-+0.10u)$.
Horizons and trajectory counts are drawn uniformly from the source sets above.
Test rollouts use the same $\gamma=0.97$ and independent reward noise
with standard deviation $0.03$; this noise is separate from the fixed
reward-table perturbation $\xi_{sa}$. Each main-evaluation query has one
MDP/log/target-policy triple and an exact dynamic-programming value.
The five test batches contain 300 queries each.

\subsubsection{Clinical}
\label{app:clinical-tasks}

Clinical evaluation uses the released ICU-Sepsis aggregate MDP
\citep{choudhary2024icusepsis}, which was constructed from MIMIC-III
records. Its 25 actions combine five intravenous-fluid levels and five
vasopressor levels. The published MDP has 716 states, including death,
survival, and an auxiliary absorbing state, with cluster centroids in a
47-dimensional normalized feature space. Our evaluation further aggregates
these states and generates new simulated logs. It is therefore
semi-synthetic evaluation, not a new clinical data collection or a test of
treatment effectiveness. The source prior below is a separately synthesized
treatment-like process; it does not perturb or resample the ICU-Sepsis
transition table.

\paragraph{Source states and action responses.}
Each source MDP contains 48 live states $s=0,\ldots,47$ and three absorbing
states denoting death, survival, and another terminal outcome. For action
$a\in\{0,\ldots,24\}$, define normalized dose levels
$f_a=\lfloor a/5\rfloor/4$, $v_a=(a\bmod5)/4$, and
$t_a=(f_a+v_a)/2$. The synthetic severity rank is $x_s=s/47$.
To obtain a separate geometric coordinate $y_s$, sort 48 draws from
$\mathcal N(0,\sigma_y^2)$, with $\sigma_y\sim\mathrm U(.82,1.18)$,
add an evenly spaced sequence from $-.06$ to $.06$, and standardize
to mean zero and population variance one. The latent location is
$\ell_s=(y_s,z_{s1},z_{s2})$, where each coordinate $z_{sj}$ is sampled
from $\mathcal N(0,\sigma_j^2)$ and
$\sigma_j\sim\mathrm U(.65,1.10)$ independently.

Four response mechanisms specify an action score $g_{sa}$:
\begin{equation*}
\begin{aligned}
 g_{sa}^{\rm mono}&=(w_f f_a+w_v v_a)(.58+.72x_s),\\
 g_{sa}^{\rm bi}&=-\left\{\frac{t_a-\operatorname{clip}(b_0+b_1x_s,.05,.95)}{\omega}\right\}^{2},\\
 g_{sa}^{\rm axis}&=w_f(.35+x_s)[1+h\sin(2.3\pi x_s)]f_a\\
 &\quad+w_v(.42+x_s)[1-h\cos(1.7\pi x_s)]v_a
       +k(.4+x_s)f_av_a,\\
 g_{sa}^{\rm null}&=\xi_{s1}f_a+\xi_{s2}v_a-\xi_0f_av_a.
\end{aligned}
\end{equation*}
Here $\xi_{sj}\sim\mathcal N(0,.35^2)$ and
$\xi_0\sim\mathrm U(0,.18)$. Each row is multiplied by
$\exp(\epsilon_s)$, with
$\epsilon_s\sim\mathcal N(-\sigma_e^2/2,\sigma_e^2)$, centered across
actions, and divided by its maximum absolute centered entry, floored at
$10^{-9}$. The result, clipped to $[-1,1]$, is denoted $e_{sa}$.
Parameter ranges appear in Table~\ref{tab:clinical-dynamics}. Each listed interval denotes an independent uniform draw, except where the response mechanism is specified. Vector ranges are componentwise.
For replicate $r$ and environment index $i$ (both zero-based), the mechanism
is entry $(i+r)\bmod4$ in the order monotone, biphasic, axis-specific,
near-null. The near-null mechanism retains weak, nonzero action effects.

\paragraph{Source transitions and initial distribution.}
Write $\sigma(z)=(1+\exp[-\operatorname{clip}(z,-30,30)])^{-1}$.
Before the total-terminal-mass cap, the three terminal probabilities are
\begin{equation*}
\begin{aligned}
 \widetilde p^+_{sa}&=\operatorname{clip}
   \{b_++c_+\sigma[k_+(m_+-x_s)]+A_e e_{sa},.0005,.42\},\\
 \widetilde p^-_{sa}&=\operatorname{clip}
   \{b_-+c_-\sigma[k_-(x_s-m_-)]-\eta A_e e_{sa},.0005,.34\},\\
 \widetilde p^o_{sa}&=b_o+c_o\sigma[k_o(x_s-.5)].
\end{aligned}
\end{equation*}
Multiply all three by
$\min\{1,c_T/(\widetilde p^+_{sa}+\widetilde p^-_{sa}
+\widetilde p^o_{sa})\}$, giving $p^+,p^-,p^o$ and total mass $p^T$.
The remaining probability stays within the live states.

Define $d_{ss'}=(y_s-y_{s'})^2+\lambda\sum_{j=1}^2(z_{sj}-z_{s'j})^2$.
For each state, sample an integer neighbor count uniformly between
$\max\{7,\operatorname{round}(48q_l)\}$ and
$\min\{47,\operatorname{round}(48q_h)\}$, and retain that many nearest
states other than itself. On this set $K_s$, draw
$\epsilon_{ss'}\sim\mathcal N(0,.16^2)$ and define
\[
 B_s(s')\propto\exp(-d_{ss'}/T_K+\epsilon_{ss'}),\qquad
 W_{sa}(s')\propto B_s(s')
 \exp\!\left[\operatorname{clip}\{L_e e_{sa}(y_s-y_{s'}),-.24,.24\}\right].
\]
Both distributions are normalized over $K_s$. Draw
$h_s=\operatorname{clip}(h_0+\epsilon_s^h,.39,.51)$ with
$\epsilon_s^h\sim\mathcal N(0,\sigma_h^2)$, and set
$h_{sa}=\max\{\min[h_s,.88(1-p^T_{sa})],
\min[.28,.78(1-p^T_{sa})]\}$.
The live transition probabilities are $P(s\mid s,a)=h_{sa}$ and
$P(s'\mid s,a)=(1-p^T_{sa}-h_{sa})W_{sa}(s')$ for $s'\in K_s$.
Terminal probabilities are those above, all other entries are zero,
and each absorbing state has a self-transition. Rows are normalized
numerically. Initial mass is zero on absorbing states and is drawn as
\begin{equation*}
 \begin{aligned}
 \rho_{0,\mathrm{live}}&\sim
 \operatorname{Dirichlet}\!\left(\alpha_\rho\frac{w}{\operatorname{mean}(w)}\right),\\
 w_s&=\left[(1-\tau)e^{-(x_s-c_\rho)^2/(2b_\rho^2)}
       +\tau e^{.45y_s}\right]e^{\epsilon_s^\rho},
 \end{aligned}
\end{equation*}
where $\epsilon_s^\rho\sim\mathcal N(0,.18^2)$.

\begin{table}[t]
\centering
\caption{Clinical source dynamics parameters.}
\label{tab:clinical-dynamics}
\small
\begin{tabular}{@{}ll@{}}
\toprule
Parameter & Range \\
\midrule
$(b_+,c_+,m_+,k_+)$ & $([.004,.014],[.185,.300],[.24,.42],[7,12])$ \\
$(b_-,c_-,m_-,k_-)$ & $([.003,.014],[.105,.200],[.67,.84],[7,12])$ \\
$(b_o,c_o,k_o,c_T)$ & $([.002,.013],[.001,.018],[-1.3,1.3],[.28,.34])$ \\
$(h_0,\sigma_h,T_K,\lambda)$ & $([.425,.485],[.010,.032],[.46,1.12],[.10,.38])$ \\
$(q_l,q_h)$ & $([.55,.74],[.84,.98])$ \\
$(c_\rho,b_\rho,\tau,\alpha_\rho)$ & $([.28,.72],[.13,.34],[.02,.20],[.24,1.20])$ \\
$(A_e,L_e)$, near-null & $([.001,.0045],[.005,.022])$ \\
$(A_e,L_e)$, other mechanisms & $([.012,.030],[.060,.180])$ \\
$(\eta,\sigma_e,w_f,w_v)$ & $([.35,.85],[.12,.42],[.65,1.35],[.65,1.35])$ \\
$(b_0,b_1,\omega,k,h)$ & $([.12,.36],[.35,.72],[.22,.48],[-.48,.28],[.35,.90])$ \\
\bottomrule
\end{tabular}
\end{table}

\paragraph{Source rewards.}
Rewards combine outcome-specific and continued-live components rather
than equating $\bar r(s,a)$ with survival probability. First sample a magnitude
$M=M_0M_1$, where $\log M_0\sim\mathrm U(\log .70,\log1.65)$ and
$M_1\sim\mathrm U(.92,1.12)$, and a signed-reward indicator
$Z\sim\operatorname{Bernoulli}(.12)$.
The reward parameters in Table~\ref{tab:clinical-reward-parameters} are drawn uniformly from the listed ranges. Construct positive profiles
\[
 \begin{aligned}
 \widetilde u_s^+&=\exp\{\beta[(1-x_s)_\epsilon^\kappa
                 -\operatorname{mean}((1-x)_\epsilon^\kappa)]+\epsilon_s^+\},\\
 \widetilde u_s^-&=\exp\{\beta[x_{s,\epsilon}^\kappa
                 -\operatorname{mean}(x_\epsilon^\kappa)]+\epsilon_s^-\},
 \end{aligned}
\]
where $z_\epsilon=\operatorname{clip}(z,10^{-4},1)$ and
$\epsilon_s^\pm\sim\mathcal N(0,\sigma_r^2)$.
Normalize each profile by its mean and then clip to $[.55,1.65]$,
yielding $u_s^+,u_s^-$.
The preliminary survival, death, and other-outcome rewards are
$c^+_{sa}=Mu_s^+$, $c^-_{sa}=-M\delta u_s^-$, and
$c^o_{sa}=M\zeta\operatorname{clip}(.65+.35(1-x_s)+\epsilon_s^o,.25,1.25)$,
where $\epsilon_s^o\sim\mathcal N(0,.04^2)$,
$\zeta\sim\mathrm U(0,.025)$, and $\delta=0$ if $Z=0$ or
$\delta\sim\mathrm U(.045,.18)$ if $Z=1$.

For live successors, draw $L=M(1-c_r)c_r^{-1}U_L$ with
$U_L\sim\mathrm U(.18,.48)$ and use
\begin{equation*}
 c^{\rm live}_{sa}=L\left[
 \operatorname{clip}(.18+.82(1-x_s)+\epsilon_s^l,0,1.15)
 -b_d(.55f_a+.45v_a)-b_i f_av_a-.25Zx_s\right],
\end{equation*}
where $\epsilon_s^l\sim\mathcal N(0,.05^2)$.
When $Z=0$, negative live rewards are replaced by zero.
The corresponding state-action expected reward is
\begin{equation}
 \bar r(s,a)=p^+_{sa}c^+_{sa}+p^-_{sa}c^-_{sa}
       +p^o_{sa}c^o_{sa}+(1-p^T_{sa})c^{\rm live}_{sa}.
 \label{eq:clinical-expected-reward}
\end{equation}
Absorbing states have $\bar r(s,a)=0$.

\begin{table}[t]
\centering
\caption{Clinical source reward-profile parameters.}
\label{tab:clinical-reward-parameters}
\small
\begin{tabular}{@{}lll@{}}
\toprule
Parameter & Meaning & Range \\
\midrule
$\kappa$ & Severity-profile exponent & $[.45,3.20]$ \\
$\beta$ & Profile strength & $[.04,.48]$ \\
$\sigma_r$ & Log-profile noise SD & $[.015,.14]$ \\
$c_r$ & Terminal-reward concentration parameter & $[.82,.985]$ \\
$b_d,b_i$ & Dose and interaction burdens & $[.05,.38],\ [0,.16]$ \\
\bottomrule
\end{tabular}
\end{table}

The source prior additionally varies reward concentration at a fixed
source-defined return scale. Choose a desired uniform-policy value $v_*$ at horizon 20, a desired peak $r_*$ when applicable, and a peak-state fraction $q_*$ from Table~\ref{tab:clinical-outcome-regimes}. Each listed range denotes a uniform draw; dashes indicate that no peak constraint is imposed.
For concentrated regimes, select
$\max\{1,\operatorname{round}(48q_*)\}$ live states uniformly without
replacement. Multiply their survival-outcome rewards by $b\geq1$.
For each candidate $b$, compute the resulting expected reward
$\bar r_b$ using Eq.~\eqref{eq:clinical-expected-reward}, and multiply
all outcome and live rewards by
\begin{equation}
 a_b=\frac{v_*}{J_{20}(P,\bar r_b,\rho_0,u)},
 \qquad u(a\mid s)=1/25,
 \label{eq:clinical-source-reward-calibration}
\end{equation}
where $J_{20}$ is computed by exact dynamic programming with
$\gamma=.99$. If the peak at $b=1$ is below $r_*$, select $b$ by
56 bisection steps on $[1,256]$ so that
$\max_{s,a}a_b\bar r_b(s,a)\geq r_*$; otherwise retain $b=1$.
Diffuse regimes use $b=1$ without a peak constraint. The generator
rejects an unrealizable peak rather than modifying the value label.
The final MDP uses these scaled rewards, and all logs and labels are
computed subsequently from that MDP. This operation uses only synthetic
source quantities, not ICU-Sepsis values or evaluation labels.
The outcome regime is entry $(i+2r)\bmod5$ of the table.
Separate seeded random streams generate geometry, reward magnitude,
reward concentration, hazards, occupancy, response, and outcome-regime
parameters; the final factors remain coupled through severity and
Eq.~\eqref{eq:clinical-source-reward-calibration}.

\begin{table}[t]
\centering
\caption{Clinical source outcome regimes.}
\label{tab:clinical-outcome-regimes}
\small
\begin{tabular}{@{}lccc@{}}
\toprule
Regime & $v_*$ & $r_*$ & $q_*$ \\
\midrule
Diffuse low & $[.28,.42]$ & --- & --- \\
Diffuse ordinary & $[.60,.76]$ & --- & --- \\
Concentrated ordinary & $[.58,.72]$ & $[.52,.70]$ & $[.12,.22]$ \\
Concentrated high & $[.88,1.18]$ & $[.50,.70]$ & $[.14,.25]$ \\
Competing ordinary & $[.52,.80]$ & $[.36,.58]$ & $[.10,.20]$ \\
\bottomrule
\end{tabular}
\end{table}

\paragraph{Source policies, logs, and labels.}
Let $Q_0^u$ be the time-zero action value of the uniform policy for the
sampled MDP and horizon $H$. Center each action-value row, and let $m_Q$
be the median of the live-state row standard deviations exceeding
$10^{-10}$ (use one if this set is empty). The source policy score is
\begin{equation*}
 \begin{aligned}
 z_{sa}^{\rm src}&=\operatorname{clip}\!\left(
 \frac{Q_0^u(s,a)-\operatorname{mean}_bQ_0^u(s,b)}
      {\max(m_Q,10^{-8})},-4,4\right),\\
 \pi_{q,T,\epsilon}&=(1-\epsilon)\operatorname{softmax}(qz^{\rm src}/T)
                         +\epsilon u.
 \end{aligned}
\end{equation*}
Each of 50 training and 10 validation MDPs per replicate is used for
12 logs: $H\in\{5,10,20\}$ crossed with four behavior-quality intervals
$[-2.60,-.70]$, $[-.70,0]$, $[0,.70]$, and $[.70,2.60]$.
For each log, sample $q_b$ uniformly from its interval,
$T_b\sim\mathrm U(.66,1.02)$, and
$\epsilon_b\sim\mathrm U(.24,.58)$.
Within every environment--horizon combination, assign
$n\in\{24,48,96,160\}$ to the four behavior intervals by an independent
random permutation.
Each log is queried by five target policies: partition $[-2.80,2.80]$
into five equal subintervals and draw one $q_e$ uniformly from each.
The five queries share $T_e\sim\mathrm U(.60,1.00)$ and
$\epsilon_e\sim\mathrm U(.015,.075)$.

All trajectories have $H$ transitions, including zero-reward absorbing
steps, and use $\gamma=.99$. At every step the recorded reward is exactly
$R_t=\bar r(S_t,A_t)$, with no additional reward noise.
The five queries share the same behavior log but receive separate exact
finite-horizon DP labels $J(\mathcal M,\pi_e)$.
Thus each replicate contains 600 training and 120 validation logs,
producing 3,000/600 queries. All five training replicates use this same
source law with distinct random streams.

\paragraph{Test MDPs.}
Testing starts from the released \texttt{Sepsis/ICU-Sepsis-v2} parameters,
using the default mean-admissible-action treatment for unsupported actions.
Let $P^*,r^*,\rho_0^*$ denote its transition tensor, transition-specific
reward, and initial distribution. We replace each of its three terminal
rows by a zero-reward self-loop, and cluster the remaining published
centroids into 48 groups using K-means with 10 initializations.
The three terminal states remain separate, giving 51 states in total.
For the resulting map $c(i)$ and cluster $C_s=\{i:c(i)=s\}$, use weights
$w_i\propto10^{-3}+\rho_0^*(i)$ normalized within each live cluster.
The aggregate MDP is
\begin{equation*}
\begin{aligned}
 P(s'\mid s,a)&=\sum_{i\in C_s}w_i\sum_{j\in C_{s'}}P^*(j\mid i,a),\\
 \bar r(s,a)&=\sum_{i\in C_s}w_i\sum_jP^*(j\mid i,a)r^*(i,a,j),\\
 \rho_0(s)&=\sum_{i\in C_s}\rho_0^*(i).
\end{aligned}
\end{equation*}
Absorbing rows again have self-transitions and zero reward, and transition
rows and the initial distribution are numerically normalized.
Each of five evaluation blocks uses its own clustering seed, but all
blocks reuse the same original released parameter dataset; they are not
five patient cohorts. No additional dynamics or reward perturbation is
applied in this main comparison.

\paragraph{Test policies and shared logs.}
Test policies also use the uniform-policy $Q_0^u$, but standardize it
\emph{within each state}, rather than using the source median scale:
\begin{equation}
\begin{aligned}
 z_{sa}^{\rm test}&=\frac{Q_0^u(s,a)-\operatorname{mean}_bQ_0^u(s,b)}
                         {\max\{\operatorname{SD}_bQ_0^u(s,b),10^{-4}\}},\\
 \pi_b&=.50\operatorname{softmax}(q_bz^{\rm test}/.86)+.50u,\\
 \pi_e&=.98\operatorname{softmax}(q_ez^{\rm test}/.75)+.02u.
\end{aligned}
\label{eq:clinical-test-policies}
\end{equation}
Here $q_b\in\{-.6,.6\}$ and
$q_e\in\{-1.5,-.75,0,.75,1.5\}$.
All policies are stationary and row-normalized.
Within each block, each $H\in\{5,10,20\}$ and each $q_b$ receive eight
independently simulated logs. For each log, $n$ is drawn uniformly from
$\{24,48,96,160\}$, and the five target qualities query that same log.
Thus each block has 48 logs and 240 queries, giving 240 logs and 1,200
queries across five aggregate MDPs. Rewards are the deterministic
aggregate means $\bar r(s,a)$, not newly sampled terminal-outcome rewards;
$\gamma=.99$ and trajectories retain all $H$ steps after absorption.
Reference values are exact finite-horizon DP values in the aggregate MDP.
The true matrices and $Q_0^u$ are used only to generate policies, logs,
and reference values. InCOPE receives the generated log, target policy,
known initial distribution and task horizon/discount, with behavior
probabilities estimated from the log as in Appendix~\ref{app:incope-input-features}.

\subsection{Baseline Implementations}
\label{app:baseline-implementations}

All baselines estimate the discounted return from the same behavior log
$D_n$ and target policy $\pi_e$. The implementations below apply to the main comparison
and the trajectory-count and policy-mismatch experiments.
Whenever a method requires behavior probabilities, it uses the pooled
count estimate $\widehat\pi_b$ in Eq.~\eqref{eq:incope-empirical-inputs},
with a pseudocount of $1/2$ per action, rather than the simulator's
behavior probabilities. Define cumulative importance ratios by
\begin{equation*}
  w_{it}=\prod_{k=0}^{t}
  \frac{\pi_e(A_{ik}\mid S_{ik})}{\widehat\pi_b(A_{ik}\mid S_{ik})}.
\end{equation*}
All methods use the task's $\gamma$ and $H$ and the logged rewards without
additional return normalization.

\paragraph{WPDIS.}
We evaluate WPDIS \citep{precup2000eligibility} as
\begin{equation}
  \widehat J_{\mathrm{WPDIS}}
  =\sum_{t=0}^{H-1}\gamma^t
  \frac{\sum_{i=1}^{n}w_{it}R_{it}}
       {\max\{\sum_{i=1}^{n}w_{it},10^{-8}\}}.
  \label{eq:baseline-wpdis}
\end{equation}
The denominator floor is part of the implementation used for the reported
results; it differs from exact self-normalization when the positive weight
sum is below $10^{-8}$. The code also floors each estimated behavior
probability at $10^{-8}$, which is inactive under the reported sample
budgets and the count smoother above. Importance ratios are not clipped.

\paragraph{FQE.}
We use finite-horizon tabular fitted Q evaluation, with a separate
least-squares regression at each time step
\citep{le2019batch,wang2024finegrained}.
Let $N_{t,sa}=\sum_i\mathbf1\{S_{it}=s,A_{it}=a\}$.
Starting with $\widehat Q_H=0$, its exact tabular updates are
\begin{equation*}
  \widehat Q_t(s,a)=
  \frac{\displaystyle\sum_{i:S_{it}=s,\,A_{it}=a}
    \left[R_{it}+\gamma\sum_b\pi_e(b\mid S_{i,t+1})
    \widehat Q_{t+1}(S_{i,t+1},b)\right]}
  {\max\{N_{t,sa},1\}},
\end{equation*}
for $t=H-1,\ldots,0$. An unobserved time--state--action cell is assigned
zero. Observations are not pooled across time, and no smoothing or
regularization is applied. The estimate uses the known initial distribution:
$\widehat J_{\mathrm{FQE}}=\sum_{s,a}\rho_0(s)\pi_e(a\mid s)
\widehat Q_0(s,a)$.

\paragraph{Tabular-MIS.}
Following the time-indexed construction of \citet{yin2020asymptotically},
we estimate transitions and rewards separately at each time:
\begin{equation*}
  \begin{aligned}
  \widehat P_t(s'\mid s,a)
    &=\frac{\sum_i\mathbf1\{S_{it}=s,A_{it}=a,S_{i,t+1}=s'\}}
            {\max\{N_{t,sa},1\}},\\
  \widehat r_t(s,a)
    &=\frac{\sum_iR_{it}\mathbf1\{S_{it}=s,A_{it}=a\}}
            {\max\{N_{t,sa},1\}}.
  \end{aligned}
\end{equation*}
Unobserved cells have zero reward estimates and zero transition rows;
these rows are neither smoothed nor renormalized. Starting from the
empirical initial distribution
$\widehat p_0(s)=n^{-1}\sum_i\mathbf1\{S_{i0}=s\}$, compute
\begin{equation*}
  \begin{aligned}
    \widehat p_{t+1}(s')
      &=\sum_{s,a}\widehat p_t(s)\pi_e(a\mid s)
        \widehat P_t(s'\mid s,a),\\
    \widehat J_{\mathrm{MIS}}
      &=\sum_{t=0}^{H-1}\gamma^t\sum_{s,a}
        \widehat p_t(s)\pi_e(a\mid s)\widehat r_t(s,a).
  \end{aligned}
\end{equation*}
This estimator does not require $\widehat\pi_b$. With the same initial
distribution, this tabular construction and the FQE recursion above are
algebraically equivalent; the evaluated versions differ in their use of
empirical versus known initial distributions.

\paragraph{MB-OPE.}
Pooled model-based OPE fits a time-homogeneous empirical MDP to the full
log $D_n$. It uses the reward means $\widehat r$ in
Eq.~\eqref{eq:incope-reward-statistics} and the smoothed transitions
$\widehat P$ in Eq.~\eqref{eq:incope-empirical-inputs}, pooling observations
across all time steps. The transition pseudocount is $1/2$ in total per
state--action row, distributed uniformly over successor states.
An unobserved state--action pair has zero estimated reward and a uniform
transition row. Starting from $\widehat V_H=0$, compute
\begin{equation*}
  \begin{aligned}
    \widehat V_t(s)
      &=\sum_a\pi_e(a\mid s)\left[
        \widehat r(s,a)+\gamma\sum_{s'}\widehat P(s'\mid s,a)
        \widehat V_{t+1}(s')\right],\\
    \widehat J_{\mathrm{MB}}&=\sum_s\rho_0(s)\widehat V_0(s),
  \end{aligned}
\end{equation*}
for $t=H-1,\ldots,0$. The estimator uses the known initial distribution,
without source pretraining or importance weighting. The smoothing rule
is fixed to match InCOPE's preprocessing and is not selected using test
errors.

\paragraph{OPERA.}
OPERA \citep{nie2024opera} aggregates WPDIS, FQE, Tabular-MIS, and a
scalar tabular reward-shaping control-variate estimator described below.
Let $\mathbf v$ contain their four full-log estimates. We draw $B=100$
shared bootstrap samples, each containing $m=\lfloor n^{0.6}\rfloor$
whole trajectories sampled with replacement. Every candidate, including
its behavior-probability estimate and other fitted quantities, is refitted
on each sample, producing $\mathbf v^{*(b)}$. OPERA computes
\begin{equation*}
  \begin{aligned}
    \widehat A&=\frac{m}{nB}\sum_{b=1}^{B}
      (\mathbf v^{*(b)}-\mathbf v)(\mathbf v^{*(b)}-\mathbf v)^\top,\\
    \widehat{\boldsymbol\alpha}
      &\in\underset{\mathbf1^\top\boldsymbol\alpha=1}{\arg\min}\;
        \boldsymbol\alpha^\top\widehat A\boldsymbol\alpha,
    \qquad
    \widehat J_{\mathrm{OPERA}}
      =\widehat{\boldsymbol\alpha}^\top\mathbf v.
  \end{aligned}
\end{equation*}
Weights may be negative; no ridge penalty or weight clipping is added.
If the optimum is nonunique, an eigendecomposition selects the minimizer
with the smallest Euclidean norm.
The WPDIS candidate uses Eq.~\eqref{eq:baseline-wpdis}, including its
denominator floor, in both the full-log and bootstrap evaluations.

The fourth candidate is a scalar Shaped-PDIS adaptation of
\citet{majumdar2026reward}, with a time-indexed tabular potential
$\phi_t(s)$ and $\phi_H=0$. For trajectory $i$, define
\begin{equation*}
  \begin{aligned}
    Y_i&=\sum_{t=0}^{H-1}\gamma^t w_{it}R_{it},\\
    C_i(\phi)&=\phi_0(S_{i0})+
      \sum_{t=0}^{H-1}\gamma^t w_{it}
      [\gamma\phi_{t+1}(S_{i,t+1})-\phi_t(S_{it})].
  \end{aligned}
\end{equation*}
Whole trajectories are randomly divided into two nearly equal folds.
On each fitting fold $F$, the potential minimizes
\begin{equation*}
  \frac{1}{|F|}\sum_{i\in F}
  \bigl[Y_i-\overline Y_F-C_i(\phi)+\overline C_F(\phi)\bigr]^2
  +10^{-6}\sum_{t=0}^{H-1}\sum_s\phi_t(s)^2,
\end{equation*}
where bars denote fitting-fold averages. The fitted potential is divided
by the population-form sample SD of its controls; it is set to zero if
that SD is at most float64 machine precision. On the same fitting fold,
$\lambda=-\widehat{\operatorname{Cov}}(Y,C)/
\widehat{\operatorname{Var}}(C)$, with $\lambda=0$ when the variance is
at most $10^{-14}$. The final candidate averages $Y_i+\lambda C_i$ over
held-out trajectories in both directions. Only the potential and
coefficient are cross-fitted; behavior probabilities use the complete
current sample. The ratios are neither clipped nor self-normalized.
This specifies the scalar adaptation used in the aggregate, not the
original paper's vector-control implementation.

\paragraph{DR-PPI.}
We implement PERRY's DR-PPI point estimator \citep{mandyam2026perry}
with a tabular generator and exact model expectations. Whole trajectories
are ordered by hashes of episode identifiers and assigned alternately
to two equal folds $D^{(1)},D^{(2)}$. On each fitting fold, transitions
are pooled across time and estimated using the transition smoother in
Eq.~\eqref{eq:incope-empirical-inputs}; rewards use the observed means in
Eq.~\eqref{eq:incope-reward-statistics}, with zero for unobserved cells.
Let $\widehat V_0^{(k)}(s)$ be the target-policy value from state $s$
computed by $H$-step dynamic programming in the model fitted on
$D^{(k)}$. Define the clipped PDIS return
\begin{equation*}
  Y_i^{\mathrm{clip}}
    =\sum_{t=0}^{H-1}\gamma^t\min\{w_{it},\sqrt n\}R_{it}.
\end{equation*}
The two directed estimates and their average are
\begin{equation*}
  \begin{aligned}
    \widehat J_{k\to\ell}
      &=\sum_s\rho_0(s)\widehat V_0^{(k)}(s)
        +\frac{1}{|D^{(\ell)}|}\sum_{i\in D^{(\ell)}}
        [Y_i^{\mathrm{clip}}-\widehat V_0^{(k)}(S_{i0})],\\
    \widehat J_{\mathrm{DR\text{-}PPI}}
      &=\tfrac12(\widehat J_{1\to2}+\widehat J_{2\to1}).
  \end{aligned}
\end{equation*}
The trajectory count $n$ refers to the full dataset, not an individual fold. Only the generator is
cross-fitted: $\widehat\pi_b$ is estimated once from the full log and
reused in both directions. Dynamic programming replaces auxiliary-rollout
Monte Carlo averages; no additional environment trajectories are used.
We compare point estimates, not PERRY's confidence intervals.

\paragraph{Numerical implementation.}
Baseline calculations use float64. OPERA bootstrap draws and shaping
folds use fixed seeds derived from query identifiers. Bootstrap duplicates
are treated as separate sample occurrences. In the trajectory-count and
policy-mismatch runs, a failed shaping-potential linear solve is retried
using a scaled SVD ridge solve; an OPERA equality-constraint solver
failure is retried after positive scalar rescaling of $\widehat A$.
These retries preserve the
respective objectives and do not introduce new regularization or clipping.

\subsection{Main Evaluation Details}
\label{app:main-evaluation-details}

The main comparison uses five equally sized, separately generated test
batches within each domain, totaling 6,700 queries across the four domains.
Table~\ref{tab:main-evaluation-budgets} lists the query counts, trajectory counts,
and horizons. A query pairs a behavior log of $n$ trajectories with a
target policy. Each Clinical batch uses 48 logs from one MDP, with five
target policies per log, giving 240 distinct logs and 1,200 queries overall.
Results cover all listed trajectory counts and horizons.

\begin{table}[htbp]
    \centering
    \caption{Main-comparison query counts, trajectory counts, and horizons.}
    \label{tab:main-evaluation-budgets}
    \small
    \setlength{\tabcolsep}{4pt}
    \begin{tabular}{@{}lrrll@{}}
        \toprule
        Domain & Total queries & Queries per batch
        & Trajectories $n$ & Horizon $H$ \\
        \midrule
        GridWorld & 2,500 & 500
        & $4,8,12,16,24,32$ & $12,18,24,32$ \\
        Tumor & 1,500 & 300
        & $8,16,24,32,48$ & $25$ \\
        Dispatch & 1,500 & 300
        & $4,8,12,16,24$ & $24,32,40$ \\
        Clinical & 1,200 & 240
        & $24,48,96,160$ & $5,10,20$ \\
        \bottomrule
    \end{tabular}
\end{table}

For each baseline, we compute one MAE per test batch. For each InCOPE
configuration, its five pretrained checkpoints are evaluated on every
batch, and their individual MAEs are averaged within that batch.
Table~\ref{tab:incope-main-results} reports the mean and sample standard
deviation of the resulting five batch-level MAEs, with divisor $5-1$
in the sample variance. This SD measures test-batch variability with
checkpoints held fixed, rather than pretraining variability.

\subsection{Trajectory Count and Policy Mismatch}
\label{app:log-policy-sensitivity}

This experiment uses 50 test tasks per domain, organized into five blocks of ten. For each task, the MDP, target policy, initial distribution, discount, and horizon are held fixed across all conditions. GridWorld and Dispatch use the test generators in Appendix~\ref{app:task-priors}, with horizons sampled from $\{12,18,24,32\}$ and $\{24,32,40\}$, respectively. Tumor uses $H=25$ and Clinical uses $H=10$. Clinical tasks within a block share an aggregated environment but have separately generated policies and logs. The other domains use a separately generated environment for each task.

\paragraph{Behavior-policy construction.}
Let $u(a\mid s)=1/|\mathcal A|$. We generate eight candidate behavior policies and replace each candidate $q$ by $0.95q+0.05u$. For GridWorld and Dispatch, the unmodified candidate is sampled uniformly from
\[
  q\in\{0.7225\pi_w+0.2775u,\;
          0.55\pi_w+0.45u,\;
          0.65\pi_-+0.35u,\;
          0.90\pi_-+0.10u\},
\]
corresponding to low, moderate, severe, and extreme shift. Here $\pi_w=\operatorname{softmax}(Q_0^u/T_w)$ and $\pi_-=\operatorname{softmax}(-Q_0^u/T_-)$, with fresh temperatures $T_w\sim\mathrm U(0.7,1.5)$ and $T_-\sim\mathrm U(0.2,0.8)$ per candidate. Tumor samples uniformly from the three standard behavior-policy regimes in Table~\ref{tab:tumor-policy-pairs}. Clinical uses
\[
  q_\kappa=\operatorname{softmax}(\kappa z^{\rm test}/0.75),
  \quad \kappa\in\{-2,-1.25,-0.5,0.5,1.25,2\},
\]
with uniformly sampled $\kappa$ and the score in Eq.~\eqref{eq:clinical-test-policies}. Its target is $0.98q_\kappa+0.02u$, with $\kappa$ sampled uniformly from $\{-1.5,-0.75,0,0.75,1.5\}$. Tumor target regimes are moderate and severe with probabilities $0.55$ and $0.45$; the GridWorld and Dispatch target-regime distributions are those in Appendix~\ref{app:task-priors}.

We measure policy distance using normalized discounted target-state visitation:
\begin{align*}
  w_e(s)
  &=\frac{\sum_{t=0}^{H-1}\gamma^t
      \Pr_{\mathcal M,\pi_e}(S_t=s)}{\sum_{t=0}^{H-1}\gamma^t},
  \\
  \Delta(\pi_e,q)
  &=\sum_s w_e(s)\,
      \frac12\sum_a|\pi_e(a\mid s)-q(a\mid s)|.
\end{align*}
The candidate with the largest $\Delta$ is the behavior anchor $q_\star$. This selection uses policy distance, not estimation error. The five behavior policies are
\begin{equation*}
  \pi_b^{(\alpha)}=(1-\alpha)\pi_e+\alpha q_\star,
  \qquad \alpha\in\{0,0.25,0.5,0.75,1\}.
\end{equation*}
Consequently, $\Delta(\pi_e,\pi_b^{(\alpha)})=\alpha\Delta(\pi_e,q_\star)$ for each task. Figure~\ref{fig:incope-log-policy-shift}(b) uses $\alpha$ as the policy mismatch level, with $\alpha=0$ corresponding to the target policy and $\alpha=1$ to the selected behavior anchor. Each level therefore represents a fraction of the task-specific anchor distance; absolute distances can differ across tasks and domains. The simulator supplies $w_e$ only to construct and describe this diagnostic; it is not an input to an OPE estimator.

\paragraph{Paired datasets and evaluation.}
For each task and $\alpha$, we generate one log at the largest domain-specific size and take trajectory prefixes for the smaller sizes:
\[
\begin{array}{lc}
\text{GridWorld and Dispatch} & n\in\{4,8,16,24\},\\
\text{Tumor} & n\in\{8,16,24,48\},\\
\text{Clinical} & n\in\{24,48,96,160\}.
\end{array}
\]
Figure~\ref{fig:incope-log-policy-shift}(a) fixes $\alpha=1$. Panel~(b) fixes the second-smallest trajectory count in each domain, using $n=8,16,8,48$ for GridWorld, Tumor, Dispatch, and Clinical, respectively. Different behavior policies have separately generated logs. All estimators share each evaluation log and target policy. All six baselines use the implementations in Appendix~\ref{app:baseline-implementations}. InCOPE uses its five pretrained checkpoints without additional training.

\paragraph{Uncertainty.}
Each point is the mean absolute error over test tasks, additionally averaged over the five checkpoints for InCOPE. The shaded bands are pointwise $95\%$ percentile intervals from 2,000 bootstrap draws. Each draw resamples the five blocks and then the ten tasks within each sampled block; for InCOPE, it also independently resamples the five training replicates. Task draws are shared across methods and conditions, and replicate draws are shared by specialist and unified models. The intervals therefore include test-task variation and, for InCOPE, training-replicate variation.

\subsection{Pretraining Prior Ablations}
\label{app:prior-ablations}

The experiments in Section~\ref{sec:prior-design} use domain-specific specialists with the architecture in Appendix~\ref{app:incope-network} and training settings in Appendix~\ref{app:incope-training}. Each configuration has 3,000 source-training queries and 600 source-validation queries per domain and training replicate. We use five replicates, pairing source-task identities and training seeds across configurations. Checkpoint selection uses the corresponding source-validation distribution. Original-prior comparisons reuse the pretrained specialists; modified priors require separate training. Each experiment has a separately generated 6,700-query test set, with the domain counts and trajectory counts in Table~\ref{tab:main-evaluation-budgets}. Within an experiment, every configuration is evaluated on the same test queries.

\paragraph{Source-reward intervention.}
For domain $k$, let $\mathcal Z$ contain states whose original reward is zero for every action and whose transitions are self-loops for every action. In the randomized-reward source tasks, the reward law is
\begin{equation*}
  R_t^{\mathrm{rand}}\mid S_t=s,A_t=a
  \sim
  \begin{cases}
    \mathcal N(\mu_k,\sigma_k^2),&s\notin\mathcal Z,\\
    \delta_0,&s\in\mathcal Z,
  \end{cases}
  \qquad
  \bar r^{\mathrm{rand}}(s,a)=\mu_k\mathbf1\{s\notin\mathcal Z\}.
\end{equation*}
Here $\delta_0$ is a point mass at zero, and Gaussian draws are independent across logged transitions. We estimate $\mu_k$ and $\sigma_k$ once from nonabsorbing logged rewards in the original source-training data of the first replicate. The moment sample contains 64 equally spaced source-task groups per source provider; Dispatch has two independent providers and contributes 64 groups from each. Shared physical logs are counted once. If these sampled rewards are $r_1,\ldots,r_m$, we use
\[
  \mu_k=m^{-1}\sum_{j=1}^m r_j,
  \qquad
  \sigma_k^2=m^{-1}\sum_{j=1}^m(r_j-\mu_k)^2.
\]
These moments remain fixed across all training replicates and both source splits; they do not use test rewards or values.

The intervention retains the original transitions, initial distribution, policies, and logged states, actions, and successors. Only the logged rewards are replaced, after which all reward-dependent input features and the exact dynamic-programming value labels are recomputed. We compare source priors containing $0\%$, $50\%$, and $100\%$ original-reward tasks. The $50\%$ mixture selects complete predefined groups rather than individual queries, keeping all queries within each group together. A group contains three queries in GridWorld and Tumor, one in Dispatch, and five in Clinical. In Tumor, a group contains three separately sampled datasets with the same
trajectory count from one source MDP; the groups for the two trajectory
counts need not receive the same reward assignment. The total query count is unchanged, and the same mixture proportion applies to source training and validation.

\paragraph{Source-policy intervention.}
For the policy-range comparison, the broad setting is the original source-policy generator. The narrow setting retains the same source MDP, rewards, horizon, and trajectory count, but replaces both policies by
\begin{equation*}
  \pi_b^{\mathrm{narrow}}=0.85u+0.15\pi_b,
  \qquad
  \pi_{e,j}^{\mathrm{narrow}}=(1-\lambda_j)u+\lambda_j\pi_{e,j},
\end{equation*}
where $u$ is uniform over actions. For the three queries in a GridWorld or Tumor group, $\lambda_j\in\{0.05,0.15,0.25\}$ in query order. Dispatch uses $\lambda_1=0.15$. For the five Clinical queries, $\lambda_j\in\{0.05,0.10,0.15,0.20,0.25\}$. Logs are resampled under the modified behavior policy, value labels are recomputed for the modified target policy, and all affected input features are rebuilt. Both settings use original source rewards. The test distribution is unchanged by this intervention.

\paragraph{Aggregation.}
For a configuration and domain with $N$ test queries, let
\begin{equation}
  M_r=\frac1N\sum_{i=1}^N|\widehat J_{ri}-J_i|,
  \qquad
  \bar M=\frac15\sum_{r=1}^5M_r,
  \qquad
  s_M=\sqrt{\frac14\sum_{r=1}^5(M_r-\bar M)^2}.
  \label{eq:ablation-replicate-summary}
\end{equation}
Here $r$ indexes the training replicate, not a test batch. Table~\ref{tab:incope-prior-ablations} reports $\bar M\pm s_M$. Unlike the main-result table described in Appendix~\ref{app:main-evaluation-details}, this SD measures variation across training replicates on a fixed test set.

\subsection{Information and Estimator Controls}
\label{app:log-policy-controls}

The controls in Section~\ref{sec:behavior-log-role} use domain-specific
specialists or predictors and five training replicates. The log-free, policy-only, and full-input regression comparisons each use a separate 6,700-query test set with the domain counts and trajectory counts in Table~\ref{tab:main-evaluation-budgets}; within each comparison, all methods share the test queries. In each test set, the 1,200 Clinical queries use 240 distinct logs, with five target policies per log. Means and sample SDs are computed across replicate-level MAEs using Eq.~\eqref{eq:ablation-replicate-summary}.

\paragraph{Retraining without log-derived inputs.}
We apply the same deterministic input restriction to source training, source validation, and test queries. In the feature vector of Appendix~\ref{app:incope-input-features}, we retain the target-action probability, known initial-state mass, discount, transformed horizon, discounted horizon, and available action descriptors with their availability indicator. All other feature channels are set to zero. Thus no visit counts, reward statistics, estimated behavior probabilities, policy ratios, or log-dependent coverage and occupancy statistics remain. We replace the empirical transition tensor by
\begin{equation*}
  \widehat P_{\mathrm{LF}}(s'\mid s,a)=\frac1{|\mathcal S|},
  \qquad s,s'\in\mathcal S,\ a\in\mathcal A.
\end{equation*}
Padding masks retain the task's state and action dimensions. If the initial distribution was not externally known, its log estimate is also removed and replaced by a uniform distribution over states. The behavior-policy availability flag is set to zero. Because reward-statistic channels are zero, the reward scale $c(C_n)$ is its fixed floor of one, rather than a quantity extracted from the removed log.

The log-free model is trained from scratch with the same architecture, source-query identities, value labels, seed schedule, optimizer, and stopping rule as full InCOPE. Only the input restriction changes; the resulting stopping epochs may differ because validation errors differ. At evaluation, the same restriction is applied before prediction. This is a retrained input ablation, not masking a full-input model only at test time. Table~\ref{tab:incope-information-estimator-controls}(a) reports this comparison.

\paragraph{Policy-only regression.}
For each domain and replicate, we construct 3,000 training pairs and 600 validation pairs
\begin{equation*}
  \bigl(\operatorname{vec}(\operatorname{pad}(\pi_e)),\,
       J(\mathcal M,\pi_e)\bigr)
\end{equation*}
from the same original source tasks used for the specialists. The policy probability matrix is zero-padded to $64\times4$ for GridWorld, Tumor, and Dispatch, and to $64\times25$ for Clinical, then flattened in row-major order. No log, initial distribution, discount, horizon, action descriptors, or additional task input is supplied. Each regressor is fit separately for each domain and replicate; source-validation MAE selects among the candidates in Table~\ref{tab:policy-only-hyperparameters}. The prior-mean predictor returns the mean source-training value and has no fitted input dependence.

\begin{table}[t]
  \centering
  \caption{Settings for target-policy-only regressors.}
  \label{tab:policy-only-hyperparameters}
  \small
  \setlength{\tabcolsep}{4pt}
  \begin{tabular}{@{}lp{0.72\linewidth}@{}}
    \toprule
    Predictor & Settings \\
    \midrule
    Ridge & Training-set feature standardization; penalty $\alpha\in\{0.1,1,10,100,1000\}$; \texttt{lsqr} solver. \\
    kNN & Euclidean distance; inverse-distance weighting; $k\in\{5,25,100\}$. \\
    RBF-SVR & $C/s_y\in\{0.1,1,10\}$, $\epsilon=0.05s_y$, and kernel coefficient \texttt{gamma=scale}. \\
    ExtraTrees & 300 trees; minimum leaf size in $\{2,5,10,20\}$; $80\%$ of features considered per split. \\
    \bottomrule
  \end{tabular}
\end{table}

In Table~\ref{tab:policy-only-hyperparameters}, $s_y$ is the source-training label SD, with denominator equal to the training-set size and a floor of $10^{-8}$. For RBF-SVR, \texttt{gamma=scale} is $1/(d\,\operatorname{Var}(X))$, where $d$ is the padded input dimension and $\operatorname{Var}(X)$ is the variance over entries of the source-training input matrix. Only ridge standardizes input features; the other regressors use the padded probabilities directly. Ties in validation MAE are resolved by the smaller candidate hyperparameter. Full InCOPE uses the existing specialist checkpoints without retraining. The final policy-only comparison uses a separately generated test set evaluated after the candidate models and selection rules were fixed; test labels are not used for fitting or hyperparameter selection.

\paragraph{Policy-only results by trajectory count.}
Figure~\ref{fig:incope-policy-only-log-size} groups the existing policy-only comparison's test queries by their recorded trajectory count $n$ after evaluation. Within each domain and $n$ stratum, every method is evaluated on the same queries. We compute one MAE per method and training replicate within that stratum, then report the mean and sample SD across the five replicates. No model is retrained and no predictions are recomputed for this grouping: full InCOPE retains its complete current-task inputs, whereas the policy-only regressors still receive only the target-policy matrix. Different values of $n$ contain different test queries, rather than prefixes of the same behavior log as in Section~\ref{sec:log-policy-sensitivity}. Consequently, changes in the policy-only curves across $n$ reflect differences in query composition, not access to additional observations by those regressors. Table~\ref{tab:policy-only-overall-results} retains the overall results computed over all test queries in each domain, without stratifying by $n$.

\begin{table}[t]
    \centering
    \caption{Comparison of InCOPE with target-policy-only predictors over all test queries.}
    \label{tab:policy-only-overall-results}
    \small
    \setlength{\tabcolsep}{4pt}
    \begin{tabular}{@{}lcccc@{}}
        \toprule
        Method & GridWorld & Tumor & Dispatch & Clinical \\
        \midrule
        Full InCOPE
        & $0.782\pm0.046$
        & $0.686\pm0.192$
        & $0.317\pm0.068$
        & $0.0693\pm0.0156$ \\
        Prior mean
        & $1.093\pm0.007$
        & $3.757\pm0.026$
        & $1.897\pm0.005$
        & $0.1091\pm0.0005$ \\
        Ridge
        & $1.105\pm0.046$
        & $1.800\pm1.117$
        & $1.897\pm0.302$
        & $0.2137\pm0.0456$ \\
        kNN
        & $1.128\pm0.045$
        & $2.448\pm0.791$
        & $1.503\pm0.034$
        & $0.1143\pm0.0044$ \\
        RBF-SVR
        & $1.016\pm0.013$
        & $1.379\pm0.473$
        & $1.713\pm0.039$
        & $0.1119\pm0.0035$ \\
        ExtraTrees
        & $0.995\pm0.029$
        & $1.388\pm0.326$
        & $1.680\pm0.020$
        & $0.1126\pm0.0026$ \\
        \bottomrule
    \end{tabular}
\end{table}

\paragraph{Regression with complete task inputs.}
For each domain and training replicate, the four regressors use the same 3,000 source-training queries and exact value labels as the corresponding InCOPE specialist. The 600 source-validation queries are used only for diagnostics; the configurations below are fixed without hyperparameter search. InCOPE uses its existing specialist checkpoints. We concatenate the complete state--action feature array $X$, empirical transition tensor $\widehat P$, state and action masks, and two auxiliary availability indicators. With fixed padding to 51 states, 25 actions, and 25 features per state--action pair, this gives
\begin{equation*}
  51\cdot25\cdot25+51\cdot25\cdot51+51+51\cdot25+2
  =98{,}228
\end{equation*}
coordinates. This representation retains all information in the preprocessed inputs available to InCOPE, without feature selection or projection. It does not encode the raw trajectories beyond those inputs, and no domain identifier is added.

Ridge, kNN, and RBF-SVR use coordinatewise standardization fitted separately on each domain's source-training data for each replicate; ExtraTrees uses the unstandardized vector. Ridge uses an intercept, the \texttt{lsqr} solver, and penalty $\alpha=10$. kNN uses 25 neighbors, Euclidean distance, inverse-distance weighting, and exact nearest-neighbor search. RBF-SVR uses $C=s_y$, $\epsilon=0.05s_y$, and \texttt{gamma=scale}, where $s_y$ is the source-training label SD with population denominator and a floor of $10^{-8}$. ExtraTrees uses 100 trees, a minimum leaf size of 20, $80\%$ of features per split, the squared-error criterion, and random seed $77000+r$ for replicate $r\in\{0,\ldots,4\}$.

All methods share a new test set comprising 2,500 GridWorld, 1,500 Tumor, 1,500 Dispatch, and 1,200 Clinical queries, generated independently of the test sets used in the restricted-input comparisons. All configurations and predictions were fixed before the new target-value labels were generated. We compute one MAE per method, domain, and source-training replicate over that domain's full test set, then report the mean and sample SD across all five replicates using Eq.~\eqref{eq:ablation-replicate-summary}. No replicate is excluded and predictions are not ensembled. Table~\ref{tab:incope-information-estimator-controls}(b) reports this comparison.

\end{document}